\documentclass{article}

\usepackage[final]{neurips_2021}

\usepackage[utf8]{inputenc} 
\usepackage[T1]{fontenc}    
\usepackage{hyperref}       
\usepackage{url}            
\usepackage{booktabs}       
\usepackage{amsfonts}       
\usepackage{nicefrac}       
\usepackage{microtype}      
\usepackage{xcolor}         
\usepackage{amsmath}
\usepackage{graphicx}
\usepackage{subcaption}
\usepackage{tikz}

\setcitestyle{numbers,square}

\usepackage{bm}
\usepackage{booktabs}
\usepackage{siunitx}
\newcommand{\rpm}{\raisebox{.2ex}{$\scriptstyle\pm$}}
\newtheorem{theorem}{Theorem}
\newtheorem{lemma}{Lemma}

\title{DropGNN: Random Dropouts Increase the Expressiveness of Graph Neural Networks}

\author{
  Pál András Papp \\
  ETH Zurich \\
  \texttt{apapp@ethz.ch} \\
  \And
  Karolis Martinkus \\
  ETH Zurich \\
  \texttt{martinkus@ethz.ch} \\
  \And
  Lukas Faber \\
  ETH Zurich \\
  \texttt{lfaber@ethz.ch} \\
  \And
  Roger Wattenhofer \\
  ETH Zurich \\
  \texttt{wattenhofer@ethz.ch} \\
}

\begin{document}

\maketitle

\begin{abstract}
This paper studies Dropout Graph Neural Networks (DropGNNs), a new approach that aims to overcome the limitations of standard GNN frameworks. In DropGNNs, we execute multiple runs of a GNN on the input graph, with some of the nodes randomly and independently dropped in each of these runs. Then, we combine the results of these runs to obtain the final result. We prove that DropGNNs can distinguish various graph neighborhoods that cannot be separated by message passing GNNs. We derive theoretical bounds for the number of runs required to ensure a reliable distribution of dropouts, and we prove several properties regarding the expressive capabilities and limits of DropGNNs. We experimentally validate our theoretical findings on expressiveness. Furthermore, we show that DropGNNs perform competitively on established GNN benchmarks.
\end{abstract}

\section{Introduction}

Neural networks have been successful in handling various forms of data. Since some of the world's most interesting data is represented by graphs, Graph Neural Networks (GNNs) have achieved state-of-the-art performance in various fields such as quantum chemistry, physics, or social networks \citep{gilmer2017neural, sanchez2020learning,  kipf2017semisupervised}. On the other hand, GNNs are also known to have severe limitations and are sometimes unable to recognize even simple graph structures.

In this paper, we present a new approach to increase the expressiveness of GNNs, called Dropout Graph Neural Networks (DropGNNs). Our main idea is to execute not one but \textit{multiple} different runs of the GNN. We then aggregate the results from these different runs into a final result.

In each of these runs, we remove (``drop out'') each node in the graph with a small probability $p$. As such, the different runs of an episode will allow us to not only observe the actual 
extended neighborhood of a node for some number of layers $d$, but rather to observe various slightly perturbed versions of this $d$-hop neighborhood. We emphasize that this notion of dropouts is very different from the popular dropout regularization method; in particular, DropGNNs remove nodes during \emph{both training and testing}, since their goal is to observe a similar distribution of dropout patterns during training and testing.

This dropout technique increases the expressive power of our GNNs dramatically: even when two distinct $d$-hop neighborhoods cannot be distinguished by a standard GNN, their dropout variants (with a few nodes removed) are already separable by GNNs in most cases. Thus by learning to identify the dropout patterns where the two $d$-hop neighborhoods differ, DropGNNs can also distinguish a wide variety of cases that are beyond the theoretical limits of standard GNNs.

\subparagraph*{Our contributions.} We begin by showing several example graphs that are not distinguishable in the regular GNN setting but can be easily separated by DropGNNs. We then analyze the theoretical properties of DropGNNs in detail. We first show that executing $\widetilde{O}(\gamma)$ different runs is often already sufficient to ensure that we observe a reasonable distribution of dropouts in a neighborhood of size $\gamma$. We then discuss the theoretical capabilities and limitations of DropGNNs in general, as well as the limits of the dropout approach when combined with specific aggregation methods.

We validate our theoretical findings on established problems that are impossible to solve for standard GNNs. We find that DropGNNs clearly outperform the competition on these datasets. We further show that DropGNNs have a competitive performance on several established graph benchmarks, and they provide particularly impressive results in applications where the graph structure is really a crucial factor.

\section{Related Work}
GNNs apply deep learning to graph-structured data \citep{scarselli2008graph}. In GNNs, every node has an embedding that is shared over multiple iterations with its neighbors. This way nodes can gather their neighbors' features. In recent years, many different models have been proposed to realize how the information between nodes is shared~\citep{wu2020comprehensive}. Some approaches take inspiration from convolution~\citep{niepert2016learning, defferrard2016convolutional, hamilton2017inductive}, others from graph spectra~\citep{kipf2017semisupervised, bruna2014spectral}, others from attention~\citep{velickovic2018graph}, and others extend previous ideas of established concepts such as skip connections~\citep{xu2018jumping}.

Principally, GNNs are limited in their expressiveness by the \textit{Weisfeiler-Lehman test} (WL-test)~\citep{GIN}, a heuristic to the graph isomorphism problem. The work of \cite{GIN} proposes a new architecture, \textit{Graph Isomoprhism Networks} (GIN), that is proven to be exactly as powerful as the WL-test. However, even GINs cannot distinguish certain different graphs, namely those that the WL-test cannot distinguish. This finding~\citep{limits} motivated more expressive GNN architectures. These improvements follow two main paths.

The first approach augments the features of nodes or edges by additional information to make nodes with similar neighborhoods distinguishable. Several kinds of information have been used: inspired from distributed computing are port numbers on edges~\citep{ports}, unique IDs for nodes~\citep{loukas2020graph}, or random features on nodes~\citep{randomFeatures1, randomFeatures2}. Another idea is to use angles between edges~\citep{angles} from chemistry (where edges correspond to electron bonds).

However, all of these approaches have some shortcomings. For ports and angles, there are some simple example graphs that still cannot be distinguished with these extensions~\citep{limits}. Adding IDs or random features helps during training, but the learned models do not generalize: GNNs often tend to overfit to the specific random values in the training set, and as such, they produce weaker results on unseen test graphs that received different random values. In contrast to this, DropGNNs observe a similar distribution of embeddings during training and testing, and hence they also generalize well to test set graphs.

The second approach exploits the fact that running the WL-test on tuples, triples, or generally $k$-tuples keeps increasing its expressiveness. Thus a GNN operating on tuples of nodes has higher expressiveness than a standard GNN~\citep{morris2019weisfeiler, maron2019provably}. However, the downside of this approach is that even building a second-order graph blows up the graph quadratically. The computational cost quickly becomes a problem that needs to be to addressed, for example with sampling~\citep{morris2019weisfeiler}. Furthermore, second-order graph creation is a global operation of the graph that destroys the local semantics induced by the edges. In contrast to this, DropGNN can reason about graphs beyond the WL-test with only a small overhead (through run repetition), while also keeping the local graph structure intact.

Our work is also somewhat similar to the randomized smoothing approach \citep{cohen2019certified}, which has also been extended to GNNs recently \citep{bojchevski2020efficient}. This approach also conducts multiple runs on slightly perturbed variants of the data. However, in randomized smoothing, the different embeddings are combined in a smoothing operation (e.g. majority voting), which specifically aims to get rid of the atypical perturbed variants in order to increase robustness. In contrast to this, the main idea of DropGNNs is exactly to find and identify these perturbed special cases which are notably different from the original neighborhood, since these allow us to distinguish graphs that otherwise seem identical.

Finally, we note that removing nodes is a common tool for regularization in deep neural networks, which has also seen use in GNNs~\citep{rong2019dropedge, Grand}. However, as mentioned before, this is a different dropout concept where nodes are only removed during training to reduce the co-dependence of nodes.

\section{DropGNN}

\subsection{About GNNs}
Almost all GNN architectures~\citep{velickovic2018graph, kipf2017semisupervised, GIN, defferrard2016convolutional, wu2020comprehensive, hamilton2017inductive, xu2018jumping} follow the message passing framework~\citep{gilmer2017neural, battaglia2018relational}. Every node starts with an embedding given by its initial features. One round of message passing has three steps. In the first \textsc{message} step, nodes create a message based on their embedding and send this message to all neighbors. Second, nodes \textsc{aggregate} all messages they receive. Third, every node \textsc{update}s its embedding based on its old embedding and the aggregated messages. One such round corresponds to one GNN layer. Usually, a GNN performs $d$ rounds of message passing for some small constant $d$. Thus, the node's embedding in a GNN reflects its features and the information within its $d$-hop neighborhood. Finally, a \textsc{readout} method translates these final embeddings into predictions. Usually, \textsc{message}, \textsc{aggregate}, \textsc{update} and \textsc{readout} are functions with learnable parameters, for instance linear layers with activation functions.

This GNN paradigm is closely related to the WL-test for a pair of graphs, which is an iterative color refinement procedure. In rounds $1, ..., d$, each node looks at its own color and the multiset of colors of its direct neighbors, and uses a hash function to select a new color based on this information. As such, if the WL-test cannot distinguish two graphs, then a standard GNN cannot distinguish them either: intuitively, the nodes in these graphs receive the same messages and create the same embedding in each round, and thus they always arrive at the same final result.

\subsection{Idea and motivation}
\begin{figure}
\centering
\resizebox{0.9\textwidth}{!}{\definecolor{dgray}{gray}{0.35}

\begin{tikzpicture}

	\draw (250pt,100pt) -- (225pt,115pt);
	\draw (250pt,100pt) -- (225pt,85pt);
	\draw (250pt,100pt) -- (275pt,115pt);
	\draw (285pt,90pt) -- (275pt,115pt);
	\draw (225pt,115pt) -- (200pt,115pt);
	\draw (300pt,115pt) -- (275pt,115pt);
	\draw (225pt,85pt) -- (225pt,115pt);

	\draw[black, fill=white] (250pt,100pt) circle (6pt);
	\draw[black, fill=white] (225pt,115pt) circle (5pt);
	\draw[black, fill=white] (225pt,85pt) circle (5pt);
	\draw[black, fill=white] (200pt,115pt) circle (5pt);
	\draw[black, fill=white] (275pt,115pt) circle (5pt);
	\draw[black, fill=white] (300pt,115pt) circle (5pt);
	\draw[black, fill=white] (285pt,90pt) circle (5pt);
	
	\node[anchor=center] at (250pt,100pt) {\normalsize $u$};
	
	\draw[ultra thick, gray, arrows=-latex] (220pt,126pt) -- (205pt,135pt) -- (170pt,135pt);
	
	\draw[ultra thick, gray, arrows=-latex] (280pt,126pt) -- (295pt,135pt) -- (330pt,135pt);
	
	\draw[ultra thick, gray, arrows=-latex] (220pt,75pt) -- (205pt,66pt) -- (170pt,66pt);
	
	\draw[ultra thick, gray, arrows=-latex] (280pt,75pt) -- (295pt,66pt) -- (330pt,66pt);
	
	\draw[dgray, very thick, densely dotted] (187pt,126pt) -- (313pt,126pt) -- (313pt,75pt) -- (187pt,75pt) -- cycle;
	
	
	\draw (105pt,65pt) -- (80pt,80pt);
	\draw (105pt,65pt) -- (80pt,50pt);
	\draw (105pt,65pt) -- (130pt,80pt);
	\draw (140pt,55pt) -- (130pt,80pt);
	\draw (80pt,80pt) -- (55pt,80pt);
	\draw (155pt,80pt) -- (130pt,80pt);
	\draw (80pt,50pt) -- (80pt,80pt);

	\draw[black, fill=white] (105pt,65pt) circle (6pt);
	\draw[black, fill=white] (80pt,80pt) circle (5pt);
	\draw[black, fill=white] (80pt,50pt) circle (5pt);
	\draw[black, fill=white] (55pt,80pt) circle (5pt);
	\draw[black, fill=gray] (130pt,80pt) circle (5pt);
	\draw[black, fill=white] (155pt,80pt) circle (5pt);
	\draw[black, fill=white] (140pt,55pt) circle (5pt);
	
	\draw[ultra thick] (124pt,86pt) -- (136pt,74pt);
	\draw[ultra thick] (136pt,86pt) -- (124pt,74pt);
	
	\node[anchor=center] at (105pt,65pt) {\normalsize $u$};
	
	
	\draw (105pt,135pt) -- (80pt,150pt);
	\draw (105pt,135pt) -- (80pt,120pt);
	\draw (105pt,135pt) -- (130pt,150pt);
	\draw (140pt,125pt) -- (130pt,150pt);
	\draw (80pt,150pt) -- (55pt,150pt);
	\draw (155pt,150pt) -- (130pt,150pt);
	\draw (80pt,120pt) -- (80pt,150pt);

	\draw[black, fill=white] (105pt,135pt) circle (6pt);
	\draw[black, fill=white] (80pt,150pt) circle (5pt);
	\draw[black, fill=white] (80pt,120pt) circle (5pt);
	\draw[black, fill=white] (55pt,150pt) circle (5pt);
	\draw[black, fill=white] (130pt,150pt) circle (5pt);
	\draw[black, fill=white] (155pt,150pt) circle (5pt);
	\draw[black, fill=white] (140pt,125pt) circle (5pt);
	
	\node[anchor=center] at (105pt,135pt) {\normalsize $u$};
	
	
	\draw (390pt,65pt) -- (365pt,80pt);
	\draw (390pt,65pt) -- (365pt,50pt);
	\draw (390pt,65pt) -- (415pt,80pt);
	\draw (425pt,55pt) -- (415pt,80pt);
	\draw (365pt,80pt) -- (340pt,80pt);
	\draw (440pt,80pt) -- (415pt,80pt);
	\draw (365pt,50pt) -- (365pt,80pt);

	\draw[black, fill=white] (390pt,65pt) circle (6pt);
	\draw[black, fill=white] (365pt,80pt) circle (5pt);
	\draw[black, fill=gray] (365pt,50pt) circle (5pt);
	\draw[black, fill=white] (340pt,80pt) circle (5pt);
	\draw[black, fill=white] (415pt,80pt) circle (5pt);
	\draw[black, fill=gray] (440pt,80pt) circle (5pt);
	\draw[black, fill=white] (425pt,55pt) circle (5pt);
	
	\draw[ultra thick] (359pt,56pt) -- (371pt,44pt);
	\draw[ultra thick] (371pt,56pt) -- (359pt,44pt);
	\draw[ultra thick] (434pt,86pt) -- (446pt,74pt);
	\draw[ultra thick] (446pt,86pt) -- (434pt,74pt);
	
	\node[anchor=center] at (390pt,65pt) {\normalsize $u$};
	
	
	\draw (390pt,135pt) -- (365pt,150pt);
	\draw (390pt,135pt) -- (365pt,120pt);
	\draw (390pt,135pt) -- (415pt,150pt);
	\draw (425pt,125pt) -- (415pt,150pt);
	\draw (365pt,150pt) -- (340pt,150pt);
	\draw (440pt,150pt) -- (415pt,150pt);
	\draw (365pt,120pt) -- (365pt,150pt);

	\draw[black, fill=white] (390pt,135pt) circle (6pt);
	\draw[black, fill=white] (365pt,150pt) circle (5pt);
	\draw[black, fill=white] (365pt,120pt) circle (5pt);
	\draw[black, fill=gray] (340pt,150pt) circle (5pt);
	\draw[black, fill=white] (415pt,150pt) circle (5pt);
	\draw[black, fill=white] (440pt,150pt) circle (5pt);
	\draw[black, fill=white] (425pt,125pt) circle (5pt);
	
	\draw[ultra thick] (334pt,156pt) -- (346pt,144pt);
	\draw[ultra thick] (346pt,156pt) -- (334pt,144pt);
	
	\node[anchor=center] at (390pt,135pt) {\normalsize $u$};

\end{tikzpicture}}
\caption{Illustration of $4$ possible dropout combinations from an example $2$-hop neighborhood around $u$: a $0$-dropout, two different $1$-dropouts and a $2$-dropout.}
\label{fig:dropouts}
\end{figure}
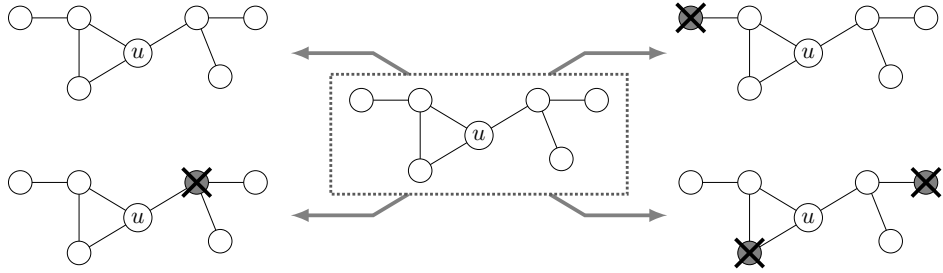

The main idea of DropGNNs is to execute multiple independent runs of the GNN during both training and testing. In each run, every node of the GNN is removed with probability $p$, independently from all other nodes. If a node $v$ is removed during a run, then $v$ does not send or receive any messages to/from its neighbors and does not affect the remaining nodes in any way. Essentially, the GNN behaves as if $v$ (and its incident edges) were not present in the graph in the specific run, and no embedding is computed for $v$ in this run (see Figure \ref{fig:dropouts} for an illustration).

Over the course of multiple runs, dropouts allow us to not only observe the $d$-hop neighborhood around any node $u$, but also several slightly perturbed variants of this $d$-hop neighborhood. In the different runs, the embedding computed for $u$ might also slightly vary, depending on which node(s) are missing from its $d$-hop neighborhood in a specific run. This increases the expressive power of GNNs significantly: even when two different $d$-hop neighborhoods cannot be distinguished by standard GNNs, the neighborhood variants observed when removing some of the nodes are usually still remarkably different. In Section \ref{sec:examples}, we discuss multiple examples for this improved expressiveness.

Our randomized approach means that in different runs, we will have different nodes dropping out of the GNN. As such, the GNN is only guaranteed to produce the same node embeddings in two runs if we have exactly the same subset of nodes dropping out. Given the $d$-hop neighborhood of a node $u$, we will refer to a specific subset of nodes dropping out as a \textit{dropout combination}, or more concretely as a $k$-dropout in case the subset has size $k$.

In order to analyze the $d$-hop neighborhood of $u$, the reasonable strategy is to use a relatively small dropout probability $p$: this ensures that in each run, only a few nodes are removed (or none at all), and thus the GNN will operate on a $d$-hop neighborhood that is similar to the original neighborhood of $u$. As a result, $1$-dropouts will be frequent, while for a larger $k$, observing a $k$-dropout will be unlikely.

To reduce the effect of randomization on the final outcome, we have to execute multiple independent runs of our GNN; we denote this number of runs by $r$. For a successful application of the dropout idea, we have to select $r$ large enough to ensure that the set of observed dropout combinations is already reasonably close to the actual probability distribution of dropouts. In practice, this will not be feasible for $k$-dropouts with large $k$ that occur very rarely, but we can already ensure for a reasonably small $r$ that e.g. the frequency of each $1$-dropout is relatively close to its expected value.

\subsection{Run aggregation}

Recall that standard GNNs first compute a final embedding for each node through $d$ layers, and then they use a \textsc{readout} method to transform this into a prediction. In DropGNNs, we also need to introduce an extra phase between these two steps, called \textit{run aggregation}.

In particular, we execute $r$ independent runs of the $d$-layer GNN (with different dropouts), which altogether produces $r$ distinct final embeddings for a node $u$. Hence we also need an extra step to merge these $r$ distinct embeddings into a single final embedding of $u$, which then acts as the input for the \textsc{readout} function. This run aggregation method has to transform a multiset of embeddings into a single embedding; furthermore, it has to be a \textit{permutation-invariant} function (similarly to neighborhood aggregation), since the ordering of different runs carries no meaning.

We note that simply applying a popular permutation-invariant function for run aggregation, such as \texttt{sum} or \texttt{max}, is often not expressive enough to extract sufficient information from the distribution of runs. Instead, one natural solution is to first apply a transformation on each node embedding, and only execute \texttt{sum} aggregation afterward. For example, a simple transformation $x \rightarrow \sigma\left( W x + b \right)$, where $\sigma$ denotes a basic non-linearity such as a sigmoid or step function, is already sufficient for almost all of our examples and theoretical results in the paper.

\subsection{Motivational examples} \label{sec:examples}

We discuss several examples to demonstrate how DropGNNs are more expressive than standard GNNs. We only outline the intuitive ideas behind the behavior of the DropGNNs here; however, in Appendix \ref{app:examples}, we also describe the concrete functions that can separate each pair of graphs.

\begin{figure}
\centering
\begin{subfigure}[b]{0.24\textwidth}
\centering
\resizebox{1.0\textwidth}{!}{\begin{tikzpicture}

	\draw (0pt,0pt) -- (40pt,0pt);
	\draw (0pt,30pt) -- (40pt,30pt);
	\draw (0pt,60pt) -- (40pt,60pt);
	\draw (0pt,90pt) -- (40pt,90pt);
	\draw (0pt,0pt) -- (0pt,30pt);
	\draw (40pt,0pt) -- (40pt,30pt);
	\draw (0pt,60pt) -- (0pt,90pt);
	\draw (40pt,60pt) -- (40pt,90pt);

	\draw[black, fill=white] (0pt,0pt) circle (6pt);
	\draw[black, fill=white] (40pt,0pt) circle (6pt);
	\draw[black, fill=white] (0pt,30pt) circle (6pt);
	\draw[black, fill=white] (40pt,30pt) circle (6pt);
	\draw[black, fill=white] (0pt,60pt) circle (6pt);
	\draw[black, fill=white] (40pt,60pt) circle (6pt);
	\draw[black, fill=white] (0pt,90pt) circle (6pt);
	\draw[black, fill=white] (40pt,90pt) circle (6pt);

	
	\draw (100pt,0pt) -- (140pt,0pt);
	\draw (100pt,90pt) -- (140pt,90pt);
	\draw (100pt,0pt) -- (100pt,90pt);
	\draw (140pt,0pt) -- (140pt,90pt);

	\draw[black, fill=white] (100pt,0pt) circle (6pt);
	\draw[black, fill=white] (100pt,30pt) circle (6pt);
	\draw[black, fill=white] (100pt,60pt) circle (6pt);
	\draw[black, fill=white] (100pt,90pt) circle (6pt);
	\draw[black, fill=white] (140pt,0pt) circle (6pt);
	\draw[black, fill=white] (140pt,30pt) circle (6pt);
	\draw[black, fill=white] (140pt,60pt) circle (6pt);
	\draw[black, fill=white] (140pt,90pt) circle (6pt);

\end{tikzpicture}}
\caption{}
\label{fig:ex_cycles}
\end{subfigure}
\hspace{0.12\textwidth}
\begin{subfigure}[b]{0.26\textwidth}
\centering
\resizebox{1.0\textwidth}{!}{\begin{tikzpicture}

	\draw (0pt,0pt) -- (40pt,0pt);
	\draw (0pt,0pt) -- (20pt,25pt);
	\draw (40pt,0pt) -- (20pt,25pt);
	\draw (20pt,25pt) -- (20pt,55pt);
	\draw (20pt,55pt) -- (0pt,80pt);
	\draw (20pt,55pt) -- (40pt,80pt);
	\draw (0pt,80pt) -- (40pt,80pt);

	\draw[black, fill=white] (0pt,0pt) circle (6pt);
	\draw[black, fill=white] (40pt,0pt) circle (6pt);
	\draw[black, fill=gray] (20pt,25pt) circle (6pt);
	\draw[black, fill=gray] (20pt,55pt) circle (6pt);
	\draw[black, fill=white] (0pt,80pt) circle (6pt);
	\draw[black, fill=white] (40pt,80pt) circle (6pt);

	
	\draw (100pt,0pt) -- (140pt,0pt);
	\draw (100pt,40pt) -- (140pt,40pt);
	\draw (100pt,80pt) -- (140pt,80pt);
	\draw (100pt,0pt) -- (100pt,80pt);
	\draw (140pt,0pt) -- (140pt,80pt);
	
	\draw[black, fill=white] (100pt,0pt) circle (6pt);
	\draw[black, fill=white] (140pt,0pt) circle (6pt);
	\draw[black, fill=gray] (100pt,40pt) circle (6pt);
	\draw[black, fill=gray] (140pt,40pt) circle (6pt);
	\draw[black, fill=white] (100pt,80pt) circle (6pt);
	\draw[black, fill=white] (140pt,80pt) circle (6pt);

\end{tikzpicture}}
\caption{}
\label{fig:ex_WL}
\end{subfigure}
\hspace{0.12\textwidth}
\begin{subfigure}[b]{0.22\textwidth}
\centering
\resizebox{1.0\textwidth}{!}{\begin{tikzpicture}

	\draw (0pt,0pt) -- (0pt,50pt);

	\draw[black, fill=white] (0pt,0pt) circle (6pt);
	\draw[black, fill=white] (0pt,25pt) circle (8pt);
	\draw[black, fill=gray] (0pt,50pt) circle (6pt);
	
	\node[anchor=center] at (0pt,25pt) {\large $u$};
	
	
	\draw (75pt,0pt) -- (75pt,50pt);
	\draw (50pt,25pt) -- (100pt,25pt);
	
	\draw[black, fill=white] (75pt,0pt) circle (6pt);
	\draw[black, fill=white] (75pt,25pt) circle (8pt);
	\draw[black, fill=gray] (75pt,50pt) circle (6pt);
	\draw[black, fill=white] (50pt,25pt) circle (6pt);
	\draw[black, fill=gray] (100pt,25pt) circle (6pt);
	
	\node[anchor=center] at (75pt,25pt) {\large $u$};

\end{tikzpicture}}
\vspace{-4pt}
\caption{}
\label{fig:ex_mean}
\end{subfigure}
\caption{Several example graphs which show that DropGNNs are more expressive than standard GNNs in various cases. Different node colors correspond to different node features.}
\label{fig:example}
\end{figure}
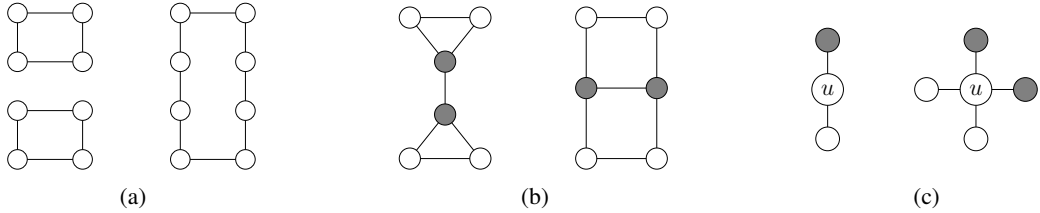

\subparagraph*{Example 1.} Figure \ref{fig:ex_cycles} shows a fundamental example of two different graphs that cannot be distinguished by the $1$-WL test, consisting of cycles of different length. This example is known to be hard for extended GNNs variants: the two cases cannot even be distinguished if we also use port numbers or angles between the edges \citep{limits}.

The simplest solution here is to consider a GNN with $d=2$ layers; this already provides a very different distribution of dropouts in the two graphs. For example, the $8$-cycle has $2$ distinct $1$-dropouts where $u$ retains both of its direct neighbors, but it only has $1$ neighbor at distance 2; such a situation is not possible in the $4$-cycle at all. Alternatively, the $4$-cycle has a $1$-dropout case with probability $p \cdot (1-p)^2$ where $u$ has $2$ direct neighbors, but no distance $2$ neighbors at all; this only happens for a $2$-dropout in the $8$-cycle, i.e. with a probability of only $p^2 \cdot (1-p)^2$. With appropriate weights, a GNN can learn to recognize these situations, and thus distinguish the two cases.

\subparagraph*{Example 2.} Figure \ref{fig:ex_WL} shows another example of two graphs that cannot be separated by a WL test; note that node features simply correspond to the degrees of the nodes. From an algorithmic perspective, it is not hard to distinguish the two graphs from specific $1$-dropout cases. Let $u$ and $v$ denote the two gray nodes in the graphs, and consider the process from $u$'s perspective. In both graphs, $u$ can recognize if $v$ is removed in a run since $u$ does not receive a ``gray'' message in the first round. However, the dropout of $v$ has a different effect in the two graphs later in the process: in the right-hand graph, it means that there is no gray neighbor at a $3$-hop distance from $u$, while in the left-hand graph, $u$ will still see a gray node (itself) in a $3$-hop distance.

Thus by identifying the $1$-dropout of $v$, an algorithm can distinguish the two graphs: if we observe runs where $u$ receives no gray message in the first round, but it receives an (aggregated) gray message in the third round, then $u$ has the left-hand neighborhood. This also means that a sufficiently powerful GNN which is equivalent to the $1$-WL test can also separate the two cases.

\subparagraph*{Example 3.} Note that using a \texttt{sum} function for neighborhood aggregation is often considered a superior choice to \texttt{mean}, since $u$ cannot separate e.g. the two cases shown in Figure \ref{fig:ex_mean} with \texttt{mean} aggregation \citep{GIN}. However, the \texttt{mean} aggregation of neighbors also has some advantages over \texttt{sum}; most notably, it the computed values do not increase with the degree of the node.

We show that dropouts also increase the expressive power of GNNs with \texttt{mean} aggregation, thus possibly making \texttt{mean} aggregation a better choice in some applications. In particular, a DropGNN with \texttt{mean} aggregation is still able to separate the two cases on Figure \ref{fig:ex_mean}.

Assume that the two colors in the figure correspond to feature values of $1$ and $-1$, and let $p=\frac{1}{4}$. In the left-hand graph, there is a $1$-dropout where $u$ ends up with a single neighbor of value $1$; hence mean aggregation yields a value of $1$ with probability $\frac{1}{4} \cdot \frac{3}{4} \approx 0.19$ in each run. However, in the right-hand graph, the only way to obtain a mean of $1$ is through a $2$-dropout or some $3$-dropouts; one can calculate that the total probability of these is only $0.06$ (see Appendix \ref{app:examples}). If we first transform all other values to $0$ (e.g. with $\sigma(x-0.5)$, where $\sigma$ is a step function), then run aggregation with \texttt{mean} or \texttt{sum} can easily separate these cases. Note that if we apply a more complex transformation at run aggregation, then separation is even much easier, since e.g. the mean value of 0.33 can only appear in the right-hand graph.

\section{Theoretical analysis} \label{sec:theory}

\subsection{Required number of runs}

We analyze DropGNNs with respect to the \textit{neighborhood of interest} around a node $u$, denoted by $\Gamma$. That is, we select a specific region around $u$, and we want to ensure that the distribution of dropout combinations in this region is reasonably close to the actual probabilities. This choice of $\Gamma$ then determines the ideal choice of $p$ and $r$ in our DropGNN.

One natural choice is to select $\Gamma$ as the entire $d$-hop neighborhood of $u$, since a GNN will always compute its final values based on this region of the graph. Note that even for this largest possible $\Gamma$, the size of this neighborhood $\gamma:=|\Gamma|$ does not necessarily scale with the entire graph. That is, input graphs in practice are often sparse, and we can e.g. assume that their node degrees are upper bounded by a constant; this is indeed realistic in many biological or chemical applications, and also a frequent assumptions in previous works \citep{ports}. In this case, having $d=O(1)$ layers implies that $\gamma$ is also essentially a constant, regardless of the size of the graph.

However, we point out that $\Gamma$ can be freely chosen as a neighborhood of any specific size. That is, even if a GNN aggregates information within a distance of $d=5$ layers, we can still select $\Gamma$ to denote, for example, only the $2$-hop neighborhood of $u$. The resulting DropGNN will still compute a final node embedding based on the entire $5$-hop neighborhood of $u$; however, our DropGNN will now only ensure that we observe a reasonable distribution of dropout combinations in the $2$-hop neighborhood of $u$.

In this sense, the size $\gamma$ is essentially a trade-off hyperparameter: while a smaller $\gamma$ will require a smaller number of runs $r$ until the distribution of dropout combinations stabilizes, a larger $\gamma$ allows us to observe more variations of the region around $u$.

\subparagraph*{1-complete dropouts.} From a strictly theoretical perspective, choosing a sufficiently large $r$ always allows us to observe every possible dropout combination. However, since the number of combinations is exponential in $\gamma$, this approach is not viable in practice (see Appendix \ref{app:Chernoff} for more details).

To reasonably limit the number of necessary runs, we focus on the so-called \emph{$1$-complete case}: we want to have enough runs to ensure that at least every $1$-dropout is observed a few times. Indeed, if we can observe each variant of $\Gamma$ where a single node is removed, then this might already allow a sophisticated algorithm to reconstruct a range of useful properties of $\Gamma$. Note that in all of our examples, a specific $1$-dropout was already sufficient to distinguish the two cases.

For any specific node $v \in \Gamma$, the probability of a $1$-dropout for $v$ is $p \cdot (1-p)^{\gamma}$ in a run (including the probability that $u$ is not dropped out). We apply the $p$ value that maximizes the probability of such a $1$-dropout; a simple differentiation shows that this maximum is obtained at $p^*=\frac{1}{1+\gamma}$.

This choice of $p$ also implies that the probability of observing a specific $1$-dropout in a run is 
\[ \frac{1}{1+\gamma} \cdot \left( \frac{\gamma}{1+ \gamma} \right)^{\gamma} \geq \frac{1}{1+\gamma} \cdot \frac{1}{e} \, . \]
Hence if we execute $r \geq e \cdot (\gamma+1) = \Omega(\gamma)$ runs, then the expected number of times we observe a specific $1$-dropout (let us denote this by $\mathbb{E}_1$) is at least $\mathbb{E}_1 \geq r \cdot \frac{1}{e} \cdot \frac{1}{1+\gamma} \geq 1$.

Moreover, one can use a Chernoff bound to show that after $ \Omega(\gamma \log \gamma)$ runs, the frequency of each $1$-dropout is sharply concentrated around $\mathbb{E}_1$. This also implies that we indeed observe each $1$-dropout at least once with high probability.

For a more formal statement, let us consider a constant $\delta \in [0,1]$ and an error probability $\frac{1}{t}<1$. Also, given a node $v \in \Gamma$ (or subset $S \subseteq \Gamma$), let $X_v$ (or $X_S$) denote the number of times this $1$-dropout ($|S|$-dropout) occurs during our runs.

\begin{theorem} \label{th:chernoff1}
If $r \geq \Omega \left( \gamma \log \gamma t \right)$, then with a probability of $1-\frac{1}{t}$, it holds that  for each $v \in \Gamma$, we have $X_v \in [\, (1\!-\!\delta) \cdot \mathbb{E}_{1\,}, \, (1\!+\!\delta) \cdot \mathbb{E}_1\, ]$.
\end{theorem}

With slightly more runs, we can even ensure that each $k$-dropout for $k \geq 2$ happens less frequently than $1$-dropouts. In this case, it already becomes possible to distinguish $1$-dropouts from multiple-dropout cases based on their frequency.

\begin{theorem} \label{th:chernoff2}
If $r \geq \Omega \left( \gamma^2 + \gamma \log \gamma t \right)$, then with a probability of $1-\frac{1}{t}$ it holds that 
\begin{itemize}
\setlength\itemsep{0.8pt}
 \item for each $v \in \Gamma$, we have $X_v \in [\,(1\!-\!\delta) \cdot \mathbb{E}_{1\,} , \, (1\!+\!\delta) \cdot \mathbb{E}_1\, ]$,
 \item for each $S \subseteq \Gamma$ with $|S| \geq 2$, we have $X_S < (1\!-\!\delta) \cdot \mathbb{E}_1$.
\end{itemize}
\end{theorem}

Since the number of all dropout combinations is in the magnitude of $2^{\gamma}$, proving this bound is slightly more technical. We discuss the proofs of these theorems in Appendix \ref{app:Chernoff}.

Note that in sparse graphs, where $\gamma$ is essentially a constant, the number of runs described in Theorems \ref{th:chernoff1} and \ref{th:chernoff2} is also essentially a constant; as such, DropGNNs only impose a relatively small (constant factor) overhead in this case.

Finally, note that these theorems only consider the dropout distribution around a specific node $u$. To ensure the same properties for all $n$ nodes in the graph simultaneously, we need to add a further factor of $n$ within the logarithm to the number of necessary runs in Theorems \ref{th:chernoff1} and \ref{th:chernoff2}. However, while this is only a logarithmic dependence on $n$, it might still be undesired in practice.

\subsection{Expressive power of DropGNNs} \label{sec:power}

In Section \ref{sec:examples}, we have seen that DropGNNs often succeed when a WL-test fails. It is natural to wonder about the capabilities and limits of the dropout approach in general; we study this question for multiple neighborhood aggregation methods separately.

We consider neighborhood aggregation with \texttt{sum} and \texttt{mean} in more detail; the proofs of the corresponding claims are discussed in Appendices \ref{app:express} and \ref{app:aggregate}, respectively. Appendix \ref{app:aggregate} also discusses briefly why \texttt{max} aggregation does not combine well with the dropout approach in practice.

\subparagraph*{Aggregation with \texttt{sum}.}

Previous work has already shown that \texttt{sum} neighborhood aggregation allows for an injective GNN design, which computes a different embedding for any two neighborhoods whenever they are not equivalent for the WL-test \citep{GIN}. Intuitively speaking, this means that \texttt{sum} aggregation has the same expressive power as a general-purpose $d$-hop distributed algorithm in the corresponding model, i.e. without IDs or port numbers. Hence to understand the expressiveness of DropGNNs in this case, one needs to analyze which embeddings can be computed by such a distributed algorithm from a specific (observed) distribution of dropout combinations.

It is already non-trivial to find two distinct neighborhoods that cannot be distinguished in the $1$-complete case. However, such an example exists, even if we also consider $2$-dropouts. That is, one can construct a pair of $d$-hop neighborhoods that are non-isomorphic, and yet they produce the exact same distribution of $1$- and $2$-dropout neighborhoods in a $d$-layer DropGNN.

\begin{theorem} \label{th:counter}
There exists a pair of neighborhoods that cannot be distinguished by $1$- and $2$-dropouts.
\end{theorem}

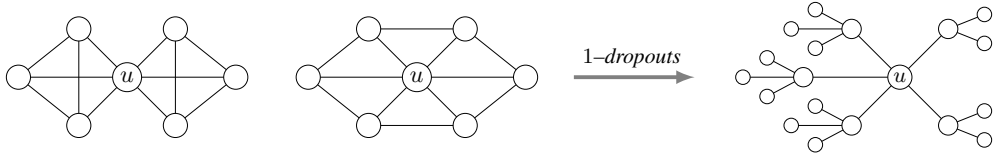
\begin{figure}
\centering
\resizebox{0.95\textwidth}{!}{\definecolor{dgray}{gray}{0.35}

\begin{tikzpicture}

	\draw (20pt,20pt) -- (45pt,0pt);
	\draw (20pt,20pt) -- (45pt,40pt);
	\draw (45pt,0pt) -- (45pt,40pt);
	\draw (110pt,20pt) -- (85pt,0pt);
	\draw (110pt,20pt) -- (85pt,40pt);
	\draw (85pt,0pt) -- (85pt,40pt);
	\draw (65pt,20pt) -- (20pt,20pt);
	\draw (65pt,20pt) -- (45pt,0pt);
	\draw (65pt,20pt) -- (45pt,40pt);
	\draw (65pt,20pt) -- (110pt,20pt);
	\draw (65pt,20pt) -- (85pt,0pt);
	\draw (65pt,20pt) -- (85pt,40pt);

	\draw[black, fill=white] (20pt,20pt) circle (5pt);
	\draw[black, fill=white] (45pt,0pt) circle (5pt);
	\draw[black, fill=white] (45pt,40pt) circle (5pt);
	\draw[black, fill=white] (65pt,20pt) circle (6pt);
	\draw[black, fill=white] (85pt,0pt) circle (5pt);
	\draw[black, fill=white] (85pt,40pt) circle (5pt);
	\draw[black, fill=white] (110pt,20pt) circle (5pt);
	
	\node[anchor=center] at (65pt,20pt) {\normalsize $u$};
	
	
	\draw (140pt,20pt) -- (165pt,0pt);
	\draw (140pt,20pt) -- (165pt,40pt);
	\draw (165pt,0pt) -- (205pt,0pt);
	\draw (230pt,20pt) -- (205pt,0pt);
	\draw (230pt,20pt) -- (205pt,40pt);
	\draw (165pt,40pt) -- (205pt,40pt);
	\draw (185pt,20pt) -- (140pt,20pt);
	\draw (185pt,20pt) -- (165pt,0pt);
	\draw (185pt,20pt) -- (165pt,40pt);
	\draw (185pt,20pt) -- (230pt,20pt);
	\draw (185pt,20pt) -- (205pt,0pt);
	\draw (185pt,20pt) -- (205pt,40pt);

	\draw[black, fill=white] (140pt,20pt) circle (5pt);
	\draw[black, fill=white] (165pt,0pt) circle (5pt);
	\draw[black, fill=white] (165pt,40pt) circle (5pt);
	\draw[black, fill=white] (185pt,20pt) circle (6pt);
	\draw[black, fill=white] (205pt,0pt) circle (5pt);
	\draw[black, fill=white] (205pt,40pt) circle (5pt);
	\draw[black, fill=white] (230pt,20pt) circle (5pt);
	
	\node[anchor=center] at (185pt,20pt) {\normalsize $u$};
	
	\draw[ultra thick, gray, arrows=-latex] (250pt,20pt) -- (300pt,20pt);
	\node[anchor=center] at (274pt,28pt) {\footnotesize $1$\textit{--dropouts}};
		
	
	\draw (385pt,20pt) -- (345pt,20pt);
	\draw (385pt,20pt) -- (365pt,0pt);
	\draw (385pt,20pt) -- (365pt,40pt);
	\draw (385pt,20pt) -- (405pt,0pt);
	\draw (385pt,20pt) -- (405pt,40pt);
	
	\draw (345pt,20pt) -- (320pt,20pt);
	\draw (345pt,20pt) -- (330pt,28pt);
	\draw (345pt,20pt) -- (330pt,12pt);
	
	\draw (365pt,0pt) -- (340pt,0pt);
	\draw (365pt,0pt) -- (350pt,8pt);
	\draw (365pt,0pt) -- (350pt,-8pt);
	
	\draw (365pt,40pt) -- (340pt,40pt);
	\draw (365pt,40pt) -- (350pt,48pt);
	\draw (365pt,40pt) -- (350pt,32pt);
	
	\draw (405pt,0pt) -- (420pt,6pt);
	\draw (405pt,0pt) -- (420pt,-6pt);
	
	\draw (405pt,40pt) -- (420pt,46pt);
	\draw (405pt,40pt) -- (420pt,34pt);

	\draw[black, fill=white] (345pt,20pt) circle (4pt);
	\draw[black, fill=white] (365pt,0pt) circle (4pt);
	\draw[black, fill=white] (365pt,40pt) circle (4pt);
	\draw[black, fill=white] (385pt,20pt) circle (5pt);
	\draw[black, fill=white] (405pt,0pt) circle (4pt);
	\draw[black, fill=white] (405pt,40pt) circle (4pt);
	
	\draw[black, fill=white] (320pt,20pt) circle (3pt);
	\draw[black, fill=white] (330pt,28pt) circle (3pt);
	\draw[black, fill=white] (330pt,12pt) circle (3pt);
	
	\draw[black, fill=white] (340pt,40pt) circle (3pt);
	\draw[black, fill=white] (350pt,48pt) circle (3pt);
	\draw[black, fill=white] (350pt,32pt) circle (3pt);
	
	\draw[black, fill=white] (340pt,0pt) circle (3pt);
	\draw[black, fill=white] (350pt,8pt) circle (3pt);
	\draw[black, fill=white] (350pt,-8pt) circle (3pt);
	
	\draw[black, fill=white] (420pt,6pt) circle (3pt);
	\draw[black, fill=white] (420pt,-6pt) circle (3pt);
	
	\draw[black, fill=white] (420pt,46pt) circle (3pt);
	\draw[black, fill=white] (420pt,34pt) circle (3pt);
	
	\node[anchor=center] at (385pt,20pt) {\small $u$};
	
\end{tikzpicture}}
\caption{Example of two graphs not separable by $1$-dropouts (left side). In both of the graphs, for any of the $1$-dropouts, $u$ observes the same tree structure for $d=2$, shown on the right side.}
\label{fig:counterexample}
\end{figure}

We illustrate a simpler example for only $1$-dropouts in Figure \ref{fig:counterexample}. For a construction that also covers the case of $2$-dropouts, the analysis is more technical; we defer this to Appendix \ref{app:express}.

We note that even these more difficult examples can be distinguished with our dropout approach, based on their $k$-dropouts for larger $k$ values. However, this requires an even higher number of runs: we need to ensure that we can observe a reliable distribution even for these many-node dropouts.

On the other hand, our dropout approach becomes even more powerful if we combine it e.g. with the extension by port numbers introduced in \citep{ports}. Intuitively speaking, port numbers allow an algorithm to determine all paths to the removed node in a $1$-dropout, which in turn allows us to reconstruct the entire $d$-hop neighborhood of $u$. As such, in this case, $1$-complete dropouts already allow us to distinguish any two neighborhoods.

\begin{theorem} \label{th:ports}
In the setting of Theorem \ref{th:chernoff1}, a DropGNN with port numbers can distinguish any two non-isomorphic $d$-hop neighborhoods.
\end{theorem}

Finally, we note that the expressive power of DropGNNs in the $1$-complete case is closely related to the \textit{graph reconstruction problem}, which is a major open problem in theoretical computer science since the 1940s \citep{reconstruction}. We discuss the differences between the two settings in Appendix \ref{app:express}.

\subparagraph*{Aggregation with \texttt{mean}.} We have seen in Section \ref{sec:examples} that even with \texttt{mean} aggregation, DropGNNs can sometimes distinguish $1$-hop neighborhoods (that is, multisets $S_1$ and $S_2$ of features) which look identical to a standard GNN. One can also prove in general that a similar separation is possible in various cases, e.g. whenever the two multisets have the same size.

\begin{lemma} \label{th:mean}
Let $S_1 \neq S_2$ be two multisets of feature vectors with $|S_1| = |S_2|$. Then $S_1$ and $S_2$ can be distinguished by a DropGNN with \texttt{mean} neighborhood aggregation.
\end{lemma}

However, in the general case, \texttt{mean} aggregation does not allow us to separate any two multisets based on $1$-dropouts. In particular, in Appendix \ref{app:express}, we also describe an example of multisets $S_1 \cap S_2 = \emptyset$ where the distribution of means obtained from $0$- and $1$-dropouts is essentially identical in $S_1$ and $S_2$. This implies that if we want to distinguish these multisets $S_1$ and $S_2$, then the best we can hope for is a more complex approach based on multiple-node dropouts.

\section{Experiments}

In all cases we extend the base GNN model to a DropGNN by running the GNN $r$ times in parallel, doing mean aggregation over the resulting $r$ node embedding copies before the graph readout step and then applying the base GNN's graph readout. Additionally, an auxiliary readout head is added to produce predictions based on each individual run. These predictions are used for an auxiliary loss term which comprises $\frac{1}{3}$ of the final loss. 
Unless stated otherwise, we set the number of runs to $m$ and choose the dropout probability to be $p=\frac{1}{m}$, where $m$ is the mean number of nodes in the graphs in the dataset. This is based on the assumption, that in the datasets we use the GNN will usually have the receptive field which covers the whole graph. We implement random node dropout by, in each run, setting all features of randomly selected nodes to $0$. See Appendix E for more details about the experimental setup and dataset statistics. The code is publicly available\footnote{\url{https://github.com/KarolisMart/DropGNN}}.

\subsection{Datasets beyond WL}
\begin{table*}[ht]
\centering
\resizebox{\textwidth}{!}{
\begin{tabular}{@{}l*{11}{S[table-format=-3.4]}@{}}
\toprule
 & \multicolumn{2}{c}{GIN} & \multicolumn{2}{c}{+Ports} & \multicolumn{2}{c}{+IDs} & \multicolumn{2}{c}{+Random feat.} & \multicolumn{2}{c}{+Dropout}\\
\cmidrule(lr){2-3} \cmidrule(lr){4-5}  \cmidrule(lr){6-7} \cmidrule(lr){8-9} \cmidrule(lr){10-11}
{Dataset} & {Train} & {Test} & {Train} & {Test} & {Train} & {Test} & {Train} & {Test} & {Train} & {Test} \\
\midrule  
{\textsc{Limits 1} \citep{limits}} & \makebox{0.50 \rpm 0.00} & \makebox{0.50 \rpm 0.00} & \makebox{0.50 \rpm 0.00} & \makebox{0.50 \rpm 0.00} & \makebox{1.00 \rpm 0.00} & \makebox{0.59 \rpm 0.19} & \makebox{0.66 \rpm 0.19} & \makebox{0.66 \rpm 0.22} & \makebox{1.00 \rpm 0.00} & \makebox{\textbf{1.00 \rpm 0.00}}\\
{\textsc{Limits 2} \citep{limits}} & \makebox{0.50 \rpm 0.00} & \makebox{0.50 \rpm 0.00} & \makebox{0.50 \rpm 0.00} & \makebox{0.50 \rpm 0.00} & \makebox{1.00 \rpm 0.00} & \makebox{0.61 \rpm 0.26} & \makebox{0.72 \rpm 0.17} & \makebox{0.64 \rpm 0.19} & \makebox{1.00 \rpm 0.00} & \makebox{\textbf{1.00 \rpm 0.00}}\\
{\textsc{$4$-cycles} \citep{loukas2020graph}} & \makebox{0.50 \rpm 0.00} & \makebox{0.50 \rpm 0.00} & \makebox{1.00 \rpm 0.01} & \makebox{0.84 \rpm 0.07} & \makebox{1.00 \rpm 0.00} & \makebox{0.58 \rpm 0.07} & \makebox{0.75 \rpm 0.05} & \makebox{0.77 \rpm 0.05} & \makebox{0.99 \rpm 0.03} & \makebox{\textbf{1.00 \rpm 0.01}}\\
{\textsc{LCC} \citep{randomFeatures1}} & \makebox{0.41 \rpm 0.09} & \makebox{0.38 \rpm 0.08} & \makebox{1.0 \rpm 0.00} & \makebox{0.39 \rpm 0.09} & \makebox{1.00 \rpm 0.00} & \makebox{0.42 \rpm 0.08} & \makebox{0.45 \rpm 0.16} & \makebox{0.46 \rpm 0.08} & \makebox{1.00 \rpm 0.00} & \makebox{\textbf{0.99 \rpm 0.02}}\\
{\textsc{Triangles} \citep{randomFeatures1}} & \makebox{0.53 \rpm 0.15} & \makebox{0.52 \rpm 0.15} & \makebox{1.0 \rpm 0.00} & \makebox{0.54 \rpm 0.11} & \makebox{1.00 \rpm 0.00} & \makebox{0.63 \rpm 0.08} & \makebox{0.57 \rpm 0.08} & \makebox{0.67 \rpm 0.05} & \makebox{0.93 \rpm 0.12} & \makebox{\textbf{0.93 \rpm 0.13}}\\
{\textsc{Skip-circles} \citep{chen2019equivalence}} & \makebox{0.10 \rpm 0.00} & \makebox{0.10 \rpm 0.00} & \makebox{1.00 \rpm 0.00} & \makebox{0.14 \rpm 0.08} & \makebox{1.00 \rpm 0.00} & \makebox{0.10 \rpm 0.09} & \makebox{0.16 \rpm 0.11} & \makebox{0.16 \rpm 0.05} & \makebox{0.81 \rpm 0.28} & \makebox{\textbf{0.82 \rpm 0.28}}\\
\bottomrule
\end{tabular}}
\caption{Evaluation of techniques that increase GNN expressiveness on challenging synthetic datasets. We highlight the best test scores in bold. Compared to other augmentation techniques DropGNN (GIN +Dropout) achieves high training accuracy but also generalizes well to the test set.}  
\label{tab:expressiveness}
\end{table*}  

To see the capabilities of DropGNN in practice we test on existing synthetic datasets, which are known to require expressiveness beyond the WL-test. We use the datasets from~\citet{randomFeatures1} that are based on $3-$regular graphs. Nodes have to predict whether they are part of a triangle (\textsc{Triangles}) or have to predict their local clustering coefficient (\textsc{LCC}). We test on the two counterexamples \textsc{Limits 1} (Figure~\ref{fig:ex_cycles}) and \textsc{Limits 2} from ~\citet{limits} where we compare two smaller structures versus one larger structure. We employ the dataset by \citet{loukas2020graph} to classify graphs on containing a cycle of length 4 (\textsc{$4$-cycles}). We increase the regularity in this dataset by ensuring that each node has a degree of 2. Finally we experiment on circular graphs with skip links (\textsc{Skip-circles}) by~\citet{chen2019equivalence}, where the model needs to classify if a given circular graph has skip links of length $\{2, 3, 4, 5, 6, 9, 11, 12, 13, 16\}$.

For comparison, we try several other GNN modifications which increase expressiveness. For control, we run a vanilla GNN on these datasets. We then extend this base GNN with (i) ports~\citep{ports} (randomly assigned), (ii) node IDs~\citep{loukas2020graph} 
(randomly permuted), and (iii) a random feature from the standard normal distribution~\citep{randomFeatures1}. The architecture for all GNNs is a 4-layer GIN with \texttt{sum} as aggregation and $\varepsilon=0$. For DropGNN $r=50$ runs are performed. For the \textsc{Skip-circles} dataset we use a 9-layer GIN instead, as the skip links can form cycles of up to 17 hops. 

We train all methods for $1,000$ epochs and then evaluate the accuracy on the training set. We then test on a new graph (with new features). We report training and testing averaged across $10$ initializations in Table~\ref{tab:expressiveness}. We can see that DropGNN outperforms the competition.

\subsection{Sensitivity analysis}
\begin{figure}
\begin{subfigure}{0.3\textwidth}
    \centering
    \includegraphics[width=\textwidth]{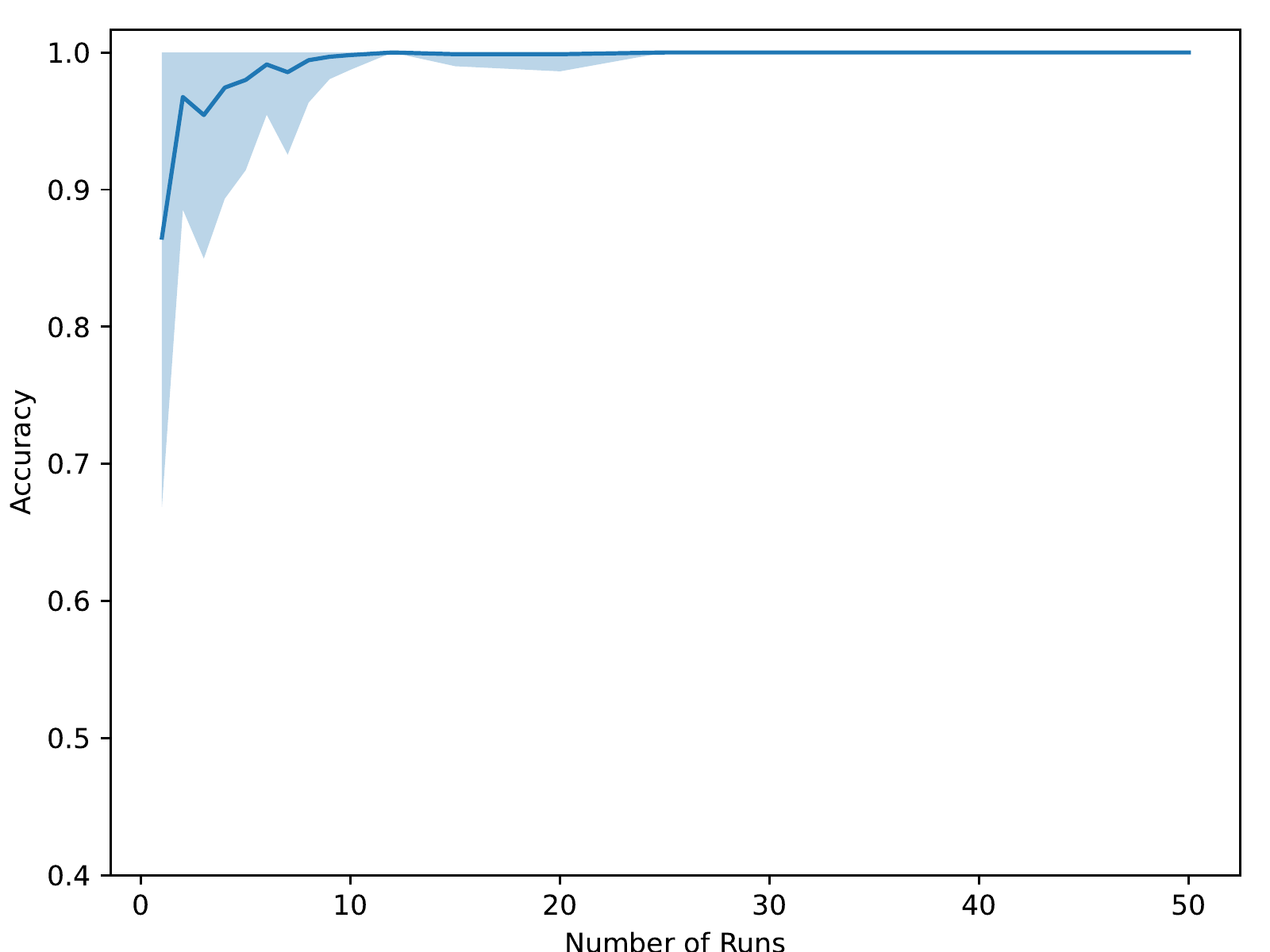}
    \caption{\textsc{Limits 1}}
    \label{fig:run_ablation_limitsone}
\end{subfigure}\hfill
\begin{subfigure}{0.3\textwidth}
    \centering
    \includegraphics[width=\textwidth]{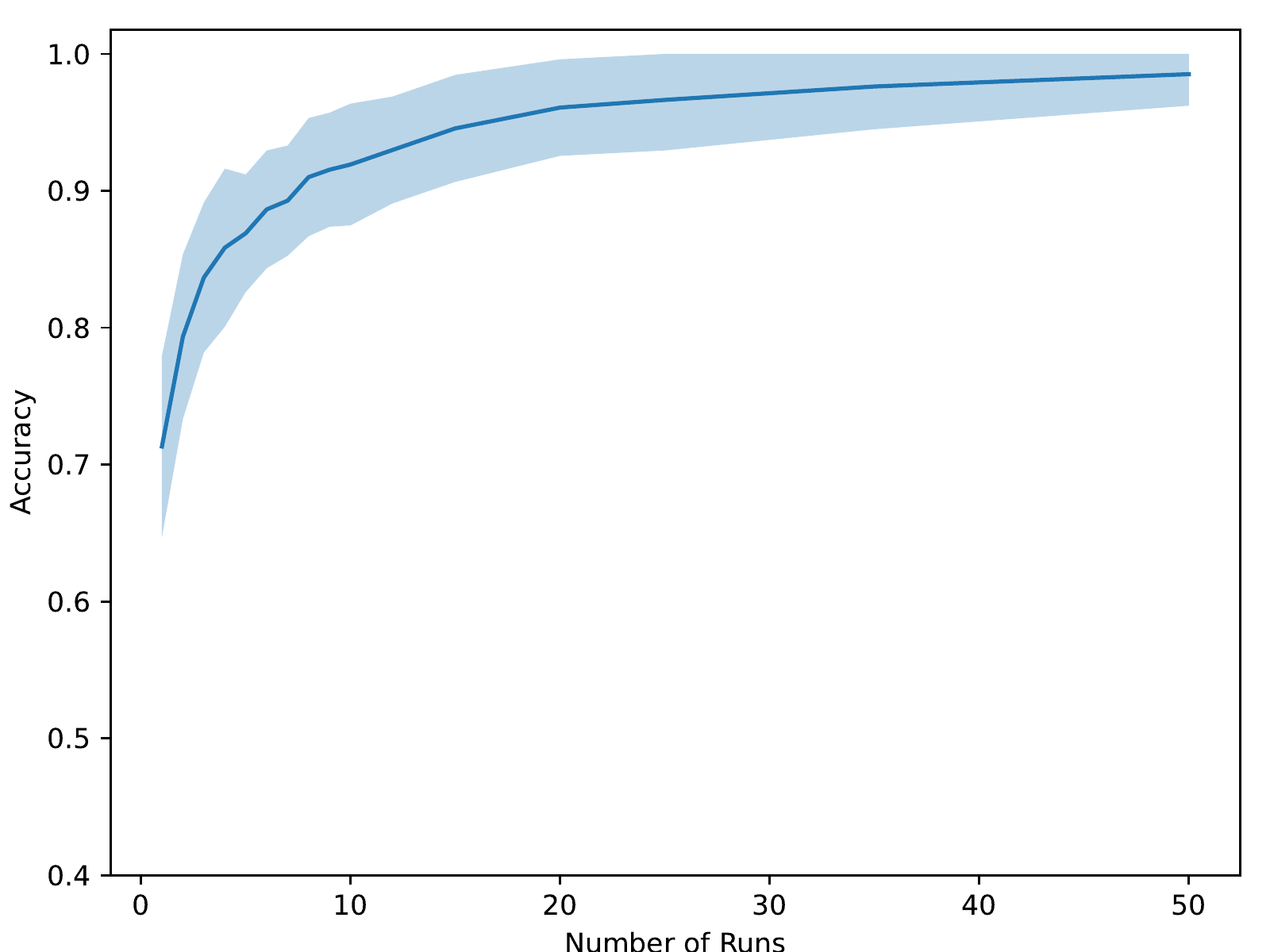}
    \caption{\textsc{$4-$cycles}}
    \label{fig:run_ablation_4cycles}
\end{subfigure}\hfill
\begin{subfigure}{0.3\textwidth}
    \centering
    \includegraphics[width=\textwidth]{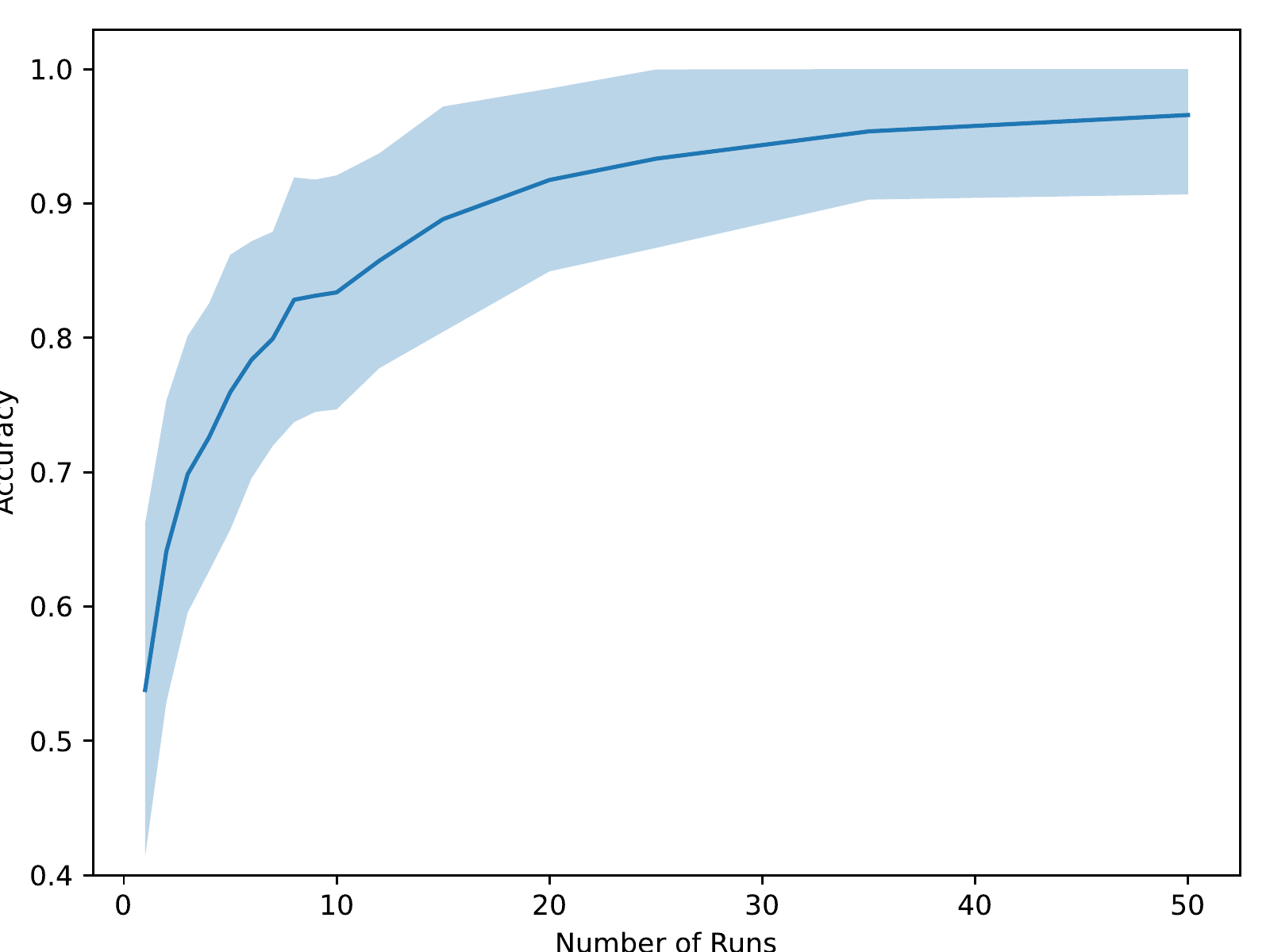}
    \caption{\textsc{Triangles}}
    \label{fig:run_ablation_triangles}
\end{subfigure}
    \caption{Investigating the impact of the number of runs ($x-$axis) versus the classification accuracy ($y-axis$). In all three plots, having more runs allows for more stable dropout observations, increasing accuracy. The tradeoff is higher runtime since the model computes more runs.}
    \label{fig:run_ablation}
\end{figure}

We investigate the impact of the number of independent runs on the overall accuracy. Generally, we expect an increasing number of runs to more reliably produce informative dropouts. We train with a sufficiently high number of runs ($50$) with the same setting as before. Now, we reevaluate DropGNN but limit the runs to a smaller number. We measure the average accuracy over $10$ seeds with $10$ tests each and plot this average in Figure~\ref{fig:run_ablation} on three datasets: \textsc{Limits 1} (Figure~\ref{fig:run_ablation_limitsone}), \textsc{$4$-cycles} (Figure~\ref{fig:run_ablation_4cycles}), and \textsc{Triangles} (Figure~\ref{fig:run_ablation_triangles}). In all three datasets, more runs directly translate to higher accuracy.

\begin{figure}
\begin{subfigure}{0.3\textwidth}
    \centering
    \includegraphics[width=\textwidth]{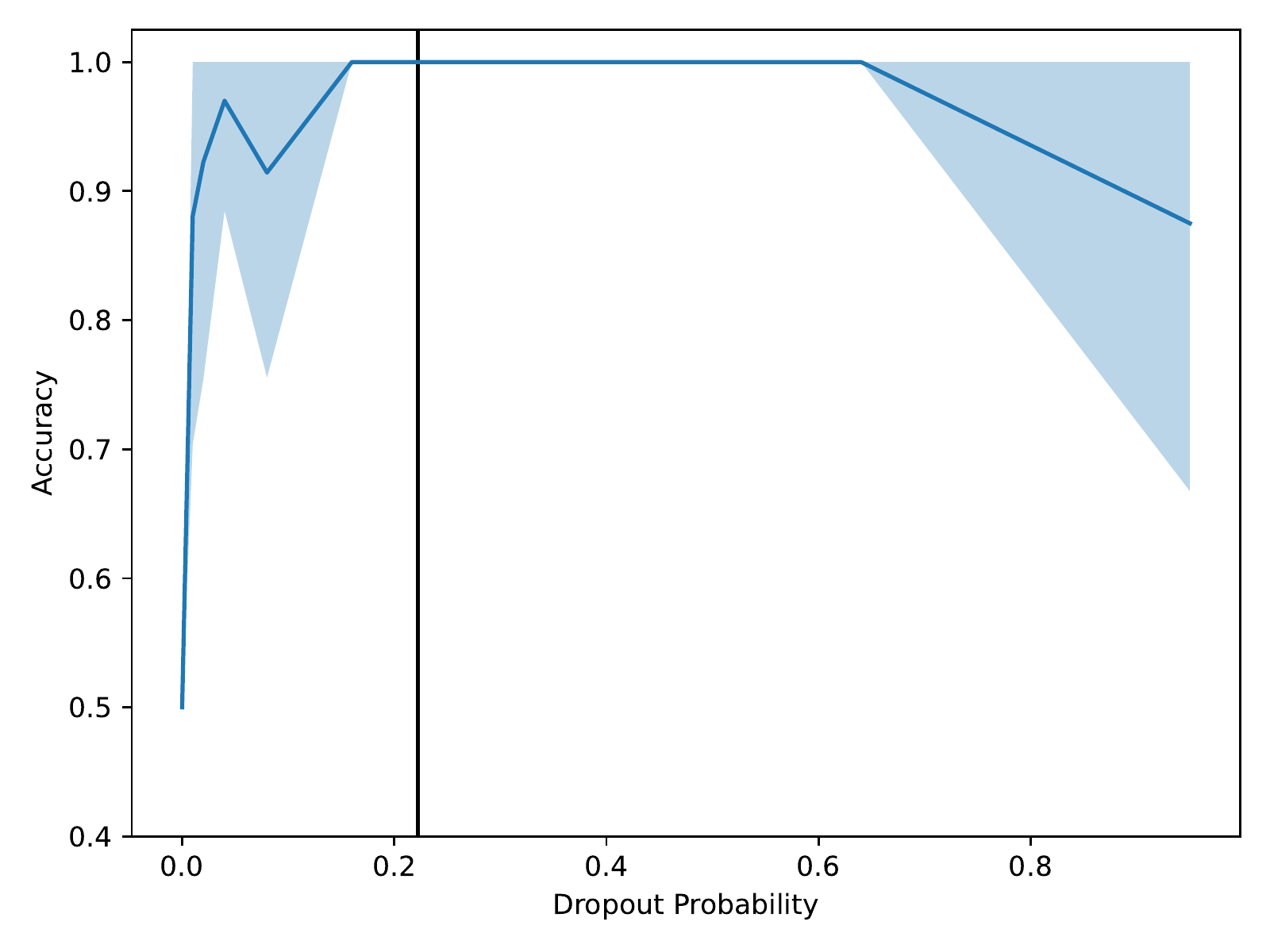}
    \caption{\textsc{Limits 1}}
    \label{fig:p_ablation_limitsone}
\end{subfigure}\hfill
\begin{subfigure}{0.3\textwidth}
    \centering
    \includegraphics[width=\textwidth]{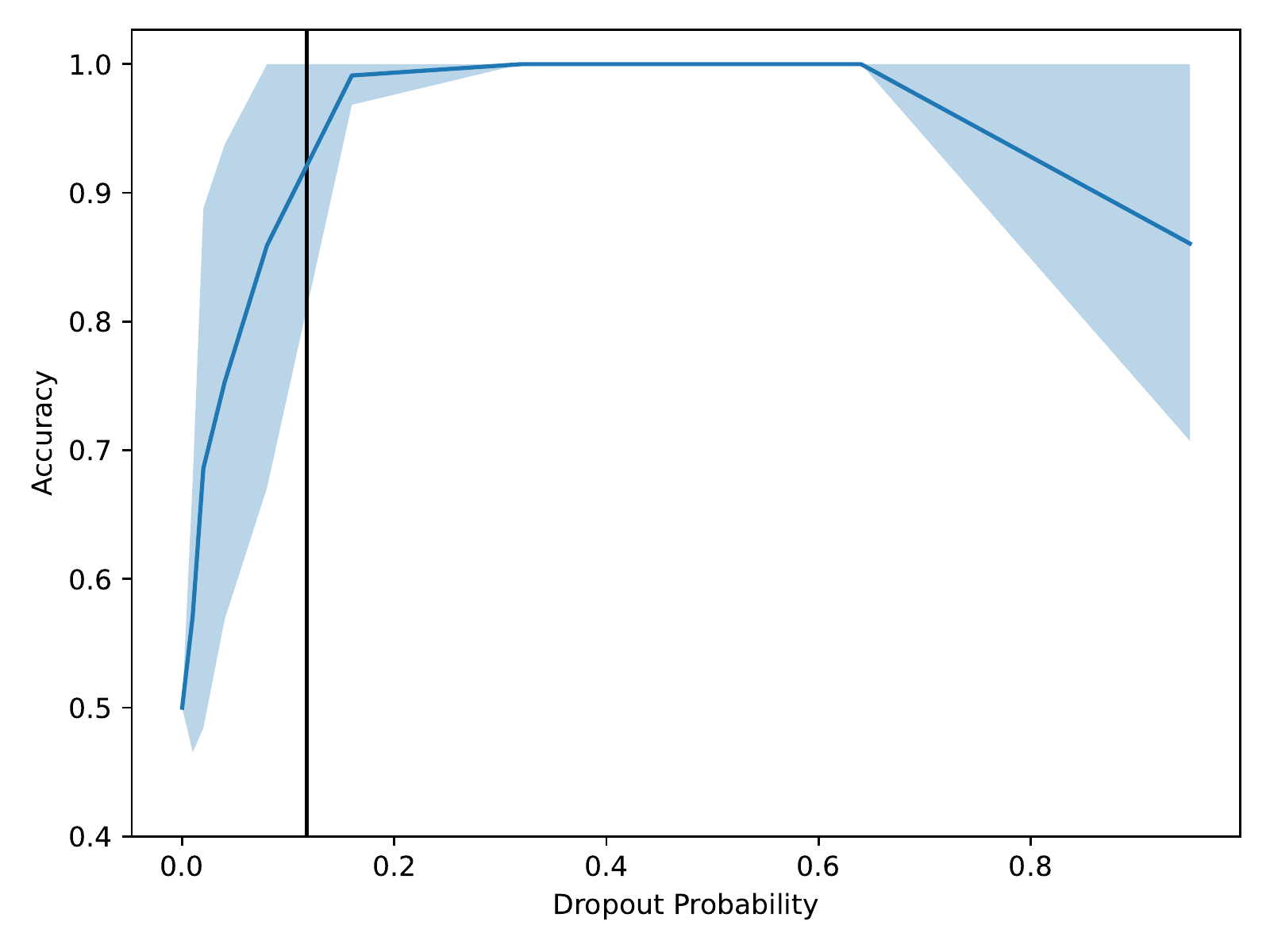}
    \caption{\textsc{$4-$cycles}}
    \label{fig:p_ablation-4cycles}
\end{subfigure}\hfill
\begin{subfigure}{0.3\textwidth}
    \centering
    \includegraphics[width=\textwidth]{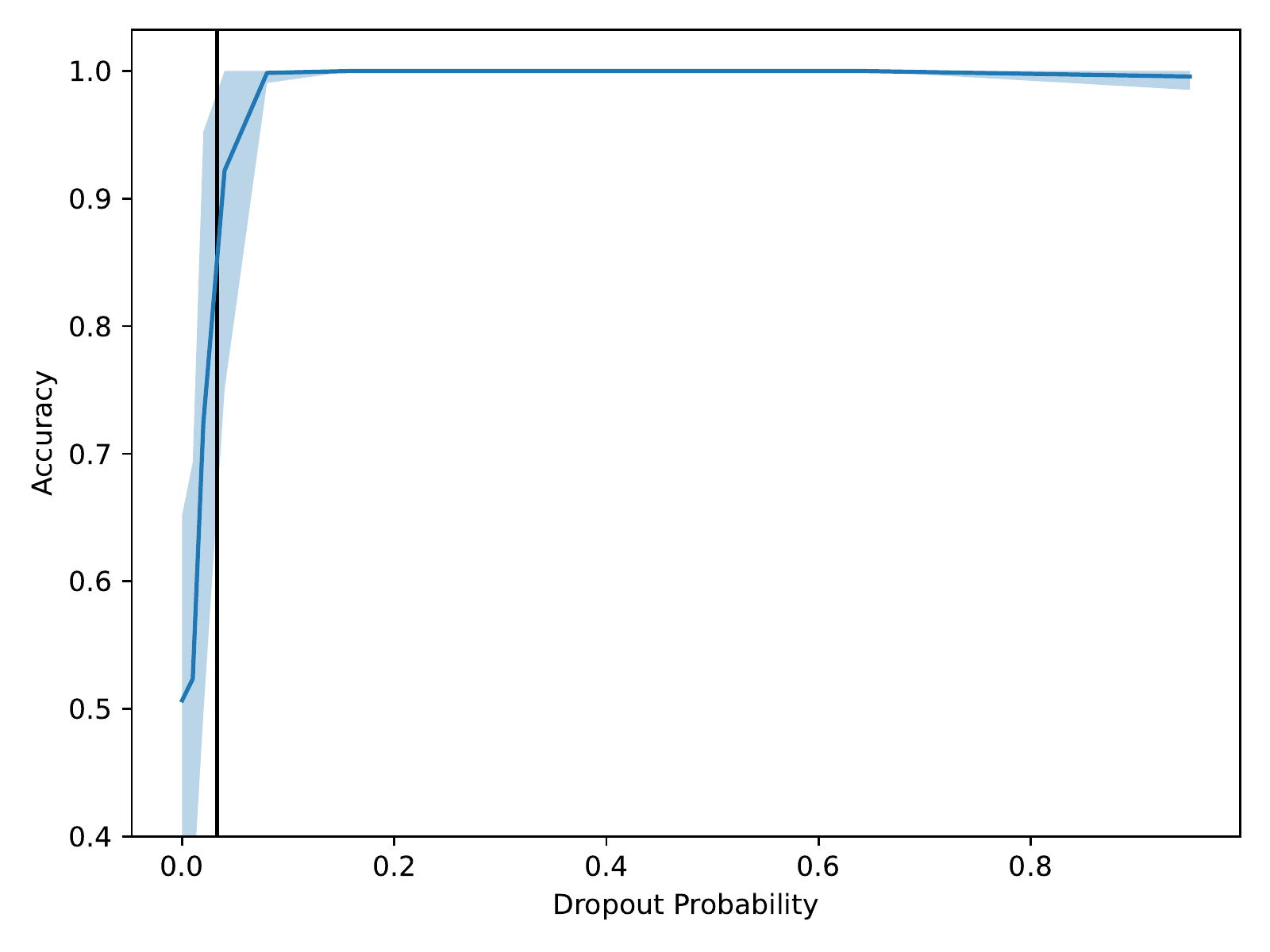}
    \caption{\textsc{Triangles}}
    \label{fig:p_ablation_triangles}
\end{subfigure}
    \caption{Investigating the impact of the dropout probability ($x-$axis) versus the classification accuracy ($y-$axis). DropGNN is robust to the choice of $p$ for decently small $p$. Choosing $p\approx \gamma^{-1}$ is a decent default that is shown by vertical black lines.}
    \label{fig:p_ablation}
\end{figure}

Next, we investigate the impact of the dropout probability $p$. We use the same setting as before, but instead of varying the number of runs in the reevaluation, we train and test with different probabilities $p$ on an exponential scale from $0.01$ to $0.64$. We also try $0$ (no dropout) and $0.95$ (almost everything is dropped). Figure~\ref{fig:p_ablation} shows the accuracy for each dropout probability, again averaged over $10$ seeds with $10$ tests each. Generally, DropGNN is robust to different values of $p$ until $p$ becomes very large.

\subsection{Graph classification}

\begin{table*}[ht]
\centering
\resizebox{0.9\textwidth}{!}{
\begin{tabular}{@{}l*{7}{S[table-format=-3.4]}@{}}
\toprule
{Model} & {Complexity} & {MUTAG} & {PTC} & {PROTEINS} & {IMDB-B} & {IMDB-M} & \\
\midrule  
{WL subtree \citep{GIN, shervashidze2011weisfeiler}} & \makebox{$O(n)$} &  \makebox{90.4 \rpm 5.7} & \makebox{59.9 \rpm 4.3} &  \makebox{75.0 \rpm 3.1} & \makebox{73.8 \rpm 3.9} & \makebox{50.9 \rpm 3.8}\\
{DCNN \citep{atwood2016diffusion}} & \makebox{$O(n)$} &  \makebox{ - } & \makebox{ - } &  \makebox{61.3 \rpm 1.6} & \makebox{49.1 \rpm 1.4} & \makebox{33.5 \rpm 1.4}\\
{PatchySan \citep{niepert2016learning}} & \makebox{$O(n)$} &  \makebox{89.0 \rpm 4.4} & \makebox{62.3 \rpm 5.7} &  \makebox{75.0 \rpm 2.5} & \makebox{71.0 \rpm 2.3} & \makebox{45.2 \rpm 2.8}\\
{DGCNN \citep{zhang2018end}} & \makebox{$O(n)$} &  \makebox{85.8 \rpm 1.7} & \makebox{58.6 \rpm 2.5} &  \makebox{75.5 \rpm 0.9} & \makebox{70.0 \rpm 0.9} & \makebox{47.8 \rpm 0.9}\\
{GIN \citep{GIN}} & \makebox{$O(n)$} &  \makebox{89.4 \rpm 5.6} & \makebox{64.6 \rpm 7.0} &  \makebox{76.2 \rpm 2.8} & \makebox{75.1 \rpm 5.1} & \makebox{\textbf{52.3 \rpm 2.8}}\\
{DropGIN (ours)} & \makebox{$O(rn), r\approx 20$} & \makebox{\textbf{90.4 \rpm 7.0}} & \makebox{\textbf{66.3 \rpm 8.6}} &  \makebox{\textbf{76.3 \rpm 6.1}} & \makebox{\textbf{75.7 \rpm 4.2}} & \makebox{51.4 \rpm 2.8}\\
\midrule
{1-2-3 GNN \citep{morris2019weisfeiler}} & \makebox{$O(n^4)$} &  \makebox{86.1} & \makebox{60.9} &  \makebox{75.5} & \makebox{\textbf{74.2}} & \makebox{49.5}\\
{PPGN \citep{maron2019provably}*}  & \makebox{$O(n^3)$} & \makebox{\textbf{90.6 \rpm 8.7}} & \makebox{\textbf{66.2 \rpm 6.5}} &  \makebox{\textbf{77.2 \rpm 4.7}} & \makebox{73 \rpm 5.8} & \makebox{\textbf{50.5 \rpm 3.6}}\\
\bottomrule
\end{tabular}}
\caption{Graph classification accuracy (\%). The best performing model in each complexity class is highlighted in bold. *We report the best result achieved by either of the three versions of their model.}  
\label{tab:graph_classification}
\end{table*}  

We evaluate and compare our modified GIN model (DropGIN) with the original GIN model and other GNN models of various expressiveness levels on real-world graph classification datasets. We use three bioinformatics datasets (MUTAG, PTC, PROTEINS) and two social networks (IMDB-BINARY and IMDB-MULTI) \citep{yanardag2015deep}. Following \citep{GIN} node degree is used as the sole input feature for the IMDB datasets, while for the bioinformatics datasets the original categorical node features are used.

We follow the evaluation and model selection protocol described in \citep{GIN} and report the 10-fold cross-validation accuracies \citep{yanardag2015deep}. We extend the original 4-layer GIN model described in \citep{GIN} and use the same hyper-parameter selection as \citep{GIN}. 
From Figure \ref{fig:p_ablation} we can see that it is usually safer to use a slightly larger $p$ than a slightly smaller one. Due to this, we set the node dropout probability to $p=\frac{2}{m}$, where $m$ is the mean number of nodes in the graphs in the dataset.

Our method successfully improves the results achieved by the original GIN model on the bioinformatics datasets (Table \ref{tab:graph_classification}) and is, in general, competitive with the more complex and computationally expensive expressive GNNs. Namely, the 1-2-3 GNN \citep{morris2019weisfeiler} which has expressive power close to that of 3-WL and $O(n^4)$ time complexity, and the Provably Powerful Graph Network (PPGN) \citep{maron2019provably} which has 3-WL expressive power and $O(n^3)$ time complexity. Compared to that, our method has only $O(rn)$ time complexity.
However, we do observe, that our approach slightly underperforms the original GIN model on the IMDB-M dataset.
Since the other expressive GNNs also underperform when compared to the original GIN model, it is possible that classifying graphs in this dataset rarely requires higher expressiveness. In such cases, our model can lose accuracy compared to the base GNN as many runs are required to achieve a fully stable dropout distribution. 

\subsection{Graph property regression}

\begin{table*}[ht]
\centering
\resizebox{0.9\textwidth}{!}{
\begin{tabular}{@{}l*{8}{S[table-format=-3.4]}@{}}
\toprule
{Property} & \makebox{Unit} & {MPNN \citep{wu2018moleculenet}} & {1-GNN \citep{morris2019weisfeiler}} & {1-2-3 GNN \cite{morris2019weisfeiler}} & {PPGN \cite{maron2019provably}} & {DropMPNN} & {Drop-1-GNN}\\
\midrule  
{$\mu$} & \makebox{Debye} &  \makebox{0.358} & \makebox{0.493} &  \makebox{0.473} & \makebox{0.0934} & \makebox{\textbf{0.059}*} & \makebox{0.453*}\\
{$\alpha$} & \makebox{$\text{Bohr}^3$} &  \makebox{0.89} & \makebox{0.78} &  \makebox{0.27} & \makebox{0.318} & \makebox{\textbf{0.173}*} & \makebox{0.767*}\\
{$\epsilon_{\text{HOMO}}$} & \makebox{Hartree} &  \makebox{0.00541} & \makebox{0.00321} &  \makebox{0.00337} & \makebox{\textbf{0.00174}} & \makebox{0.00193*} & \makebox{0.00306*}\\
{$\epsilon_{\text{LUMO}}$} & \makebox{Hartree} &  \makebox{0.00623} & \makebox{0.00350} &  \makebox{0.00351} & \makebox{0.0021} & \makebox{\textbf{0.00177}*} & \makebox{0.00306*}\\
{$\Delta\epsilon$} & \makebox{Hartree} &  \makebox{0.0066} & \makebox{0.0049} &  \makebox{0.0048} & \makebox{0.0029} & \makebox{\textbf{0.00282}*} & \makebox{0.0046*}\\
{$\langle R^2 \rangle$} & \makebox{$\text{Bohr}^2$} &  \makebox{28.5} & \makebox{34.1} &  \makebox{22.9} & \makebox{3.78} & \makebox{\textbf{0.392}*} & \makebox{30.83*}\\
{$\text{ZPVE}$} & \makebox{Hartree} &  \makebox{0.00216} & \makebox{0.00124} &  \makebox{0.00019} & \makebox{0.000399} & \makebox{\textbf{0.000112}*} & \makebox{0.000895*}\\
{$U_0$} & \makebox{Hartree} &  \makebox{2.05} & \makebox{2.32} &  \makebox{0.0427} & \makebox{\textbf{0.022}} & \makebox{0.0409*} & \makebox{1.80*}\\
{$U$} & \makebox{Hartree} &  \makebox{2.0} & \makebox{2.08} &  \makebox{0.111} & \makebox{\textbf{0.0504}} & \makebox{0.0536*} & \makebox{1.86*}\\
{$H$} & \makebox{Hartree} &  \makebox{2.02} & \makebox{2.23} &  \makebox{0.0419} & \makebox{\textbf{0.0294}} & \makebox{0.0481*} & \makebox{2.00*}\\
{$G$} & \makebox{Hartree} &  \makebox{2.02} & \makebox{1.94} &  \makebox{\textbf{0.0469}} & \makebox{0.24} & \makebox{0.0508*} & \makebox{2.12}\\ 
{$C_v$} & \makebox{cal/(mol K)} &  \makebox{0.42} & \makebox{0.27} &  \makebox{0.0944} & \makebox{\textbf{0.0144}} & \makebox{0.0596*} & \makebox{0.259*}\\

\bottomrule
\end{tabular}}
\caption{Mean absolute errors on QM9 dataset \citep{ramakrishnan2014quantum}. Best performing model is in bold and DropGNN versions that improve over the corresponding base GNN are marked with a *.}  
\label{tab:graph_regression}
\end{table*}  

We investigate how our dropout technique performs using different base GNN models on a different, graph regression, task. We use the QM9 dataset \citep{ramakrishnan2014quantum}, which consists of 134k organic molecules. The task is to predict 12 real-valued physical quantities for each molecule. In all cases, a separate model is trained to predict each quantity. We choose two GNN models to augment: MPNN~\citep{gilmer2017neural} and 1-GNN~\citep{morris2019weisfeiler}.
We set the DropGNN run count and node dropout probability the same way as done for graph classification. Following previous work \citep{morris2019weisfeiler, maron2019provably} the data is split into $80\%$ training, $10\%$ validation, and $10\%$ test sets. Both DropGNN model versions are trained for 300 epochs.

From Table~\ref{tab:graph_regression} we can see that Drop-1-GNN improves upon 1-GNN in most of the cases. In some of them, it even outperforms the much more computationally expensive 1-2-3-GNN, which uses higher-order graphs and has three times more parameters \citep{morris2019weisfeiler}. Meanwhile, DropMPNN always substantially improves on MPNN, often outperforming the Provably Powerful Graph Network (PPGN), which as you may recall scales as $O(n^3)$. This highlights the fact that while the DropGNN usually improves upon the base model, the final performance is highly dependent on the base model itself. For example, 1-GNN does not use skip connections, which might make retaining detailed information about the node's extended neighborhood much harder and this information is crucial for our dropout technique.

\section{Conclusion}\label{sec:conclusion}

We have introduced a theoretically motivated DropGNN framework, which allows us to easily increase the expressive power of existing message passing GNNs, both in theory and practice. 
DropGNNs are also competitive with more complex GNN architectures which are specially designed to have high expressive power but have high computational complexity. In contrast, our framework allows for an arbitrary trade-off between expressiveness and computational complexity by choosing the number of rounds $r$ accordingly.

\subparagraph{Societal Impact.}
In summary, we proposed a model-agnostic architecture improvement for GNNs. We do not strive to solve a particular problem but to enhance the GNN toolbox. Therefore, we do not see an immediate impact on society. We found in our experiments that DropGNN works best on graphs with smaller degrees, such as molecular graphs. Therefore, we imagine that using DropGNN in these scenarios is interesting to explore further.

\bibliography{references}
\bibliographystyle{abbrvnat}

\newpage
\appendix

\section{Concrete GNN representations for the examples} \label{app:examples}

In this section, we revisit the example graphs from Section \ref{sec:examples}, and we provide a concrete GNN implementation for each of them which is able to distinguish the two cases.

\subparagraph*{Example 1.} Let us assume for simplicity that each node starts with the integer $1$ as its single feature. Also, assume that neighborhood aggregation happens with a simple summation, with no non-linearity afterwards, and that this sum is then combined with the node's own feature again through a simple addition.

Now consider this GNN with $d=2$ layers. Note that in this case, a node $u$ in the left-hand graph is able to gather information from the whole cycle, while a node $u$ in the right-hand graph will behave as if it was the middle node in a simple path of $5$ nodes. In both cases, if no dropouts happen, then $u$ will have a value of $3$ after the first round, and a value of $9$ after the second round.

However, the $1$-dropouts are already significantly different: in the left-hand graph, they will produce a result of $5$, $5$ and $7$, while in the right-hand graph, they result in a final value of $5$, $5$, $8$ and $8$. One can similarly compute the $k$-dropouts for $k \geq 2$, which will also produce a range of other values (but at most $7$ in any case).

If we apply a more sophisticated transformation on these embeddings before run aggregation, then it is straightforward to separate these two distributions. For example, we can use an MLP to only obtain a positive value in case if the embedding is $8$ (we discuss this technique in more detail at Example $2$); this will happen regularly for the right-hand graph, but never for the left-hand graph. After this, a simple \texttt{sum} run aggregation already distinguishes the cases.

However, if one prefers a simpler transformation, then a choice of $\sigma(x-8)$ also suffices (with $\sigma$ denoting the Heaviside step function). With this transformation, a run aggregation with \texttt{sum} simply counts the number of cases when the final embedding was a $9$. Since the probability of the $0$-dropout is different in the two graphs, the expected value of this count will also differ by at least $\Omega(p \cdot r)$ after $r$ runs, which makes them straightforward to distinguish.

\subparagraph*{Example 2.} For an elegant representation of Example 2, the most convenient method is to apply a slightly more complex non-linearity for neighborhood aggregation; this allows a very simple representation for everything else in the GNN.

In particular, let us again assume that each node simply starts with an integer $1$ as a feature (i.e. not even aware of its degree initially). Furthermore, assume that neighborhood aggregation happens with a simple \texttt{sum} operator; however, after this, we use a more sophisticated non-linearity $\hat{\sigma}$ which ensures $\hat{\sigma}(2)=1$, and $\hat{\sigma}(x)=0$ for all other integers $x$. One can easily implement this function with a $2$-layer MLP: we can use $x_1=\sigma(x-1)$ and $x_2=\sigma(-x+3)$ as two nodes in the first layer, and then combine them with a single node $\sigma(x_1+x_2-1)$ as the second layer. Finally, for the \textsc{update} function which merges the aggregated neighborhood $x_{N(u)}$ with the node's own embedding $x_u$, let us select $\sigma(x_{N(u)} + x_u - 2)$.

The resulting GNN can be described rather easily. Each node begins with a feature of $1$, and has an embedding of either $0$ or $1$ in any subsequent round. The update rule for the embedding is also simple: if $u$'s own value is $1$ and $u$ has exactly $2$ neighbors with a value of $1$, then the embedding of $u$ will remain $1$; in any other case, $u$'s embedding is set to $0$, and it will remain $0$ forever.

In case of dropouts, this GNN will behave differently in the two graphs of Example $2$. Note that in both cases, whenever the connected component  containing node $u$ is not a cycle after the dropouts, then in at most $d=3$ rounds, the embedding of $u$ is set to $0$. On the other hand, if the component containing $u$ is a cycle, then the embedding of $u$ will remain $1$ after any number of rounds.

Now let $u$ denote one of the nodes with degree $3$ in both graphs. In the left-hand graph, there is a $1$-dropout (of the other gray node) that puts $u$ in a cycle, so $u$ will produce a final embedding of $1$ relatively frequently. Besides this, there are also $2$ distinct $2$-dropouts and a $3$-dropout that removes the other gray node but keeps the triangle containing $u$ intact; these will all result in a final embedding of $1$ for $u$. On the other hand, in the right-hand graph, there are only $2$ distinct $2$-dropouts which result in a single cycle containing $u$.

This means that the probability of getting a final value of $1$ is significantly higher in the left graph. In particular, after $r$ runs, the difference in the expected frequency of getting a $1$ is at least $\Omega(p \cdot r)$, so we can easily separate the two cases by executing run aggregation with \texttt{sum} or \texttt{mean}.

\subparagraph*{Example 3.} The base idea of this separation has already been outlined in Section \ref{sec:examples}: assume that the middle node $u$ uses a simple \texttt{mean} aggregation of its neighbors, and the dropout probability is $p=\frac{1}{4}$. Since we are now interested in the behavior of a specific step of \texttt{mean} aggregation, we only study the GNN for $d=1$ rounds.

With $p=\frac{1}{4}$, the left-hand graph provides the following distribution of means in a DropGNN:
\[ \text{Pr}(0)=\left( \frac{3}{4} \right)^2 \quad \text{ and } \quad \text{Pr}(1)=\text{Pr}(-1)=\frac{1}{4} \cdot \frac{3}{4} \, .\]
As such, the probability of obtaining a $1$ is about $0.19$. Note that we disregarded the case when all neighbors of $u$ are removed, but we could assume for convenience that e.g. the \texttt{mean} function also returns $0$ in this case. Furthermore, we only considered the cases when $u$ is not removed, since these are the only runs when $u$ computes an embedding at all.

On the other hand, in the right-hand graph, $u$ obtains the following distribution:
\[ \text{Pr}(0)=\left( \frac{3}{4} \right)^4+4 \cdot \left( \frac{1}{4} \right)^2 \cdot \left( \frac{3}{4} \right)^2 \quad \text{, } \quad \text{Pr}\left(\frac{1}{3}\right)=\text{Pr}\left(-\frac{1}{3}\right)=2 \cdot \frac{1}{4} \cdot \left( \frac{3}{4} \right)^3 \]
\[ \text{ and } \quad \text{Pr}(1)=\text{Pr}(-1)= \left( \frac{1}{4} \right)^2 \cdot \left( \frac{3}{4} \right)^2 + 2 \cdot \left( \frac{1}{4} \right)^3 \cdot \frac{3}{4} \,. \]
This gives a probability of about $0.06$ for the value $1$.

If we apply e.g. the transformation $x \rightarrow \sigma(x-0.5)$ on these values, then the embedding $1$ is indeed significantly more frequent in the left-hand graph. Using either \texttt{mean} or \texttt{sum} for run aggregation allows us to separate the two cases: the final embeddings in the two graphs will converge to $0.19$ and $0.06$ (both multiplied by $r$ in case of \texttt{sum}).

\subparagraph*{Alternative dropout methods.} Throughout the paper, we consider a natural and straightforward version of the dropout idea: some nodes of the graph (and their incident edges) are removed for an entire run. However, we note that there are also several alternative ways to implement this dropout approach. For example, one could remove edges instead of nodes, or one could remove nodes in an asymmetrical manner (e.g., they still receive, but do not send messages). We point out that all these examples from Section 3.4. could also be distinguished under these alternative models.

\section{Required number of runs} \label{app:Chernoff}

We now discuss the proofs of Theorems \ref{th:chernoff1} and \ref{th:chernoff2}.

Note that for any specific subset $S$ of size $k$, the probability of this $k$-dropout happening in a specific run is $p^k \cdot (1-p)^{\gamma+1-k} = \left( \frac{1}{1+\gamma} \right)^k \cdot \left( \frac{\gamma}{1+\gamma} \right)^{\gamma+1-k}$. To obtain the expected frequency $\mathbb{E} X_S$ of this $k$-dropout after $r$ runs, we simply have to multiply this expression by $r$.

Furthermore, given a constant $\delta \in [0, 1]$, a Chernoff bound shows that the probability of significantly deviating from this expected value is
\[ \text{Pr}\,\left(\, X_S \notin [\, (1\!-\!\delta) \cdot \mathbb{E} X_S, \, (1\!+\!\delta) \cdot \mathbb{E} X_S \, ] \, \right) \: \leq \: 2 \cdot e ^{- \frac{\delta^2 \cdot \mathbb{E} X_S}{3}} \, . \]

Let us consider the case of Theorem \ref{th:chernoff1} first. Since we have $\gamma$ different $1$-dropouts, we can use a union bound over these dropouts to upper bound the probability of the event that \textit{any} of the nodes $v \in \Gamma$ will have $X_v \notin [\, (1\!-\!\delta) \cdot \mathbb{E}_1, \, (1\!+\!\delta) \cdot \mathbb{E}_1 \, ]$; the probability of this event is at most
\[ 2 \cdot \gamma \cdot e ^ {- \frac{\delta^2 \cdot \mathbb{E}_1}{3}} \, . \]

If we ensure that this probability is at most $\frac{1}{t}$, then the desired property follows. Note that after taking a (natural) logarithm of both sides, this is equivalent to
\[ \log ( 2 \cdot \gamma ) - \frac{\delta^2 \cdot \mathbb{E}_1}{3} \: \leq \: - \log{t} \, , \]
and thus
\[ \mathbb{E}_1 \: \geq \: \frac{3}{\delta^2}  \cdot \log ( 2 \cdot \gamma \cdot t) \, . \]
Recall that for $\mathbb{E}_1$ we have 
\[ \mathbb{E}_1= \frac{1}{1+\gamma} \cdot \left( \frac{\gamma}{1+\gamma} \right)^{\gamma} \cdot r \geq \frac{1}{1+\gamma} \cdot \frac{1}{e} \cdot r \, . \]
Due to this lower bound, it is sufficient to ensure
\[ \frac{1}{1+\gamma} \cdot \frac{1}{e} \cdot r \: \geq \: \frac{3}{\delta^2}  \cdot \log ( 2 \cdot \gamma \cdot t) \, , \]
that is,
\[ r \: \geq \: \frac{3e}{\delta^2} \cdot (\gamma+1)  \cdot \log ( 2 \cdot \gamma \cdot t) = \Omega(\gamma \cdot \log \gamma t _{\,} ) \, . \]
This completes the proof of Theorem \ref{th:chernoff1}.

For Theorem \ref{th:chernoff2}, we also need to upper bound the probability of each dropout combination of multiple nodes. Consider $k$-dropouts for a specific $k$. In this case, we have
\[ \mathbb{E} X_S = \left( \frac{1}{1+\gamma} \right)^k \cdot \left( \frac{\gamma}{1+\gamma} \right)^{\gamma+1-k} \cdot r = \frac{1}{\gamma^{k-1}} \cdot \mathbb{E}_1 \, .\]
This implies that in order to ensure $X_S < (1\!-\!\delta) \cdot \mathbb{E}_1$ in Theorem \ref{th:chernoff2}, it is sufficient to have $X_S < (1-\delta) \cdot \gamma^{k-1} \cdot \mathbb{E} X_S$. If we want to express this as $(1 + \epsilon) \cdot \mathbb{E} X_S$ for some $\epsilon$, then we get $\epsilon = (1-\delta) \cdot \gamma^{k-1} - 1$, and thus $\epsilon = \Theta\left(\gamma^{k-1}\right)$ for appropriately chosen constants. Applying a Chernoff bound (in this case, a different variant that also allows $\epsilon>1$) then gives
\[ \text{Pr}\,\left(\, X_S \geq (1\!+\!\epsilon) \cdot \mathbb{E} X_S \, \right) \: \leq \: e ^{- \frac{\epsilon^2 \cdot \mathbb{E} X_S}{2 + \epsilon}} \, . \]
Since $\epsilon = \Theta\left(\gamma^{k-1}\right)$ and $\mathbb{E} X_S = \gamma^{-(k-1)} \cdot \mathbb{E}_1$, this is in fact
\[ e ^{- \Theta(1) \cdot \gamma^{k-1} \cdot \gamma^{-(k-1)} \cdot \mathbb{E}_1} \: = \: e ^{- \Theta(1) \cdot \mathbb{E}_1} \, . \]
Note that the number of different $k$-dropouts is ${\binom{\gamma}{k}} \leq 2^{\gamma}$, so with a union bound, we can establish this property for each $k$-dropout simultaneously; for this, we need to multiply this error probability by $2^{\gamma}$. Finally, since we want to ensure this for all $k \geq 2$, we can take a union bound over $k \in \{ 2, 3, ..., \gamma \}$, getting another multiplier of $\gamma$. Thus to obtain the second condition in Theorem \ref{th:chernoff2} with error probability $\frac{1}{t}$, we need
\[ \gamma \cdot 2^{\gamma} \cdot e ^{- \Theta(1) \cdot \mathbb{E}_1} \: \leq \: \frac{1}{t} \, . \]
After taking a logarithm and reorganization, we get
\[ \mathbb{E}_1 \: \geq \: \Theta(1) \cdot \log (2^{\gamma} \cdot \gamma \cdot t) \, . \]
With our lower bound for $\mathbb{E}_1$ and a reorganization of the right side, we can reduce this to
\[ r \: \geq \: \Theta(1) \cdot (\gamma+1) \cdot \gamma \cdot \log (2 \cdot \gamma \cdot t) \: = \: \Omega\left(\gamma^2 + \log \gamma t \right) \, . \]
Another union bound shows that the two conditions of Theorem \ref{th:chernoff2} also hold simultaneously when $r$ is in this magnitude, thus completing the proof of Theorem \ref{th:chernoff2}.

Note that if we want to ensure this property for the neighborhood of all the $n$ nodes in the graph simultaneously, then we also have to take a union bound over all the $n$ nodes, which results in a factor of $n$ within the logarithm in our final bounds on $r$.

\subparagraph*{Asymptotic analysis.} Finally, let us note that from a strictly theoretical perspective, if we consider $\gamma$ to be a constant, and $p$ to be some function of $\gamma$, then the probability of any specific $k$-dropout is $p^k \cdot (1-p)^{\gamma+1-k}$, i.e. a constant value. As such, a Chernoff bound shows that if we select $r$ to be a sufficiently large constant, then every possible dropout combination is observed, and their frequencies are reasonable close to the expected values.

However, this approach is clearly not realistic in practice: e.g. for our choice of $p \approx \gamma^{-1}$, the probability of a specific $k$-dropout is less than $p^k \approx \gamma^{-k}$. This means that we need $r \geq \gamma^k$ runs even to observe this $k$-dropout at least once in expectation. While this $\gamma^k$ is, asymptotically speaking, only a constant value, it still induces a very large overhead in practice, even for relatively small $k$ and $\gamma$ values.

\subparagraph*{Different $\gamma$ and $p$ values.} Note that our choice of $\gamma$ was defined for an arbitrary node of the graph; however, the dropout probability $p$, chosen as a function of $\gamma$, is a global parameter of DropGNNs. As such, our choice of $p$ from the analysis only works well if we assume that the graph is relatively homogeneous, i.e. $\gamma$ is similar for every node.

In practice, one can simply apply the average or the maximum of these different $\gamma$ values; a slightly smaller/larger than optimal $p$ only means that we observe some dropouts with slightly lower probability, or we execute slightly more runs than necessary. The ablation studies in Figures~\ref{fig:run_ablation} and \ref{fig:p_ablation} also show that our approach is generally robust to different number of runs and different dropout probabilities. We note, however, that if e.g. the graph consists of several different but separately homogeneous regions, then a more sophisticated approach could apply a different $p$ value in each of these regions.

\section{Expressiveness with \texttt{sum} aggregation} \label{app:express}

We now discuss our claims on DropGNNs with \texttt{sum} neighborhood aggregation. Recall that with this aggregation method, a GNN with injective functions (such as GIN) has the same expressive power as the WL-test.

Note that in this setting, we understand a $d$-hop neighborhood around $u$ to refer to the part of the graph that $u$ can observe in $d$ rounds of message passing. In particular, this contains (i) all nodes that are at most $d$ hops away from $u$, and (ii) all the edges induced by these nodes, except for the edges where both endpoints are exactly at distance $d$ from $u$.

\subsection{Proof of Theorem \ref{th:counter}}

To prove Theorem \ref{th:counter}, we show two different $d$-neighborhoods around a node $u$ (for $d=2$) that are non-isomorphic, but they generate the exact same distribution of observations for $u$ if we only consider the case of $k$-dropouts for $k \leq 2$. 

Note that the example graphs on Figure \ref{fig:counterexample} already provide an example where the $0$-dropout and the $1$-dropouts are identical. One can easily check this from the figure: in case of no dropouts, $u$ observes the same tree representation in $d=2$ steps, and in case of any of the $6$ possible $1$-dropouts (in either of the graphs), $u$ observes the tree structure shown on the right side of the figure.  

To also extend this example to the case of $2$-dropouts, we need to slightly change it. Note that the example graph is essentially constructed in the following way: we take two independent cycles of length $3$ in one case, and a single cycle of length $6$ in the other case, and in both graphs, we connect all these nodes to an extra node $u$. This construction is easy to generalize to larger cycle lengths. In particular, let us consider an integer $\ell \geq 3$, and create the following two graphs: in one of them, we take two independent cycles of length $\ell$, and connect each node to an extra node $u$, while in the other one, we take a single cycle of length $2 \cdot \ell$, and connect each node to an extra node $u$.

We claim that with a choice of $\ell = 5$, this construction suffices for Theorem \ref{th:counter}. As before, one can easily verify that $u$ observes the same $2$-hop neighborhood in case of no dropouts, and also identical $2$-hop neighborhoods for any of the $10$ possible $1$-dropouts in both graphs. The latter essentially has the same structure as the right-hand tree in Figure \ref{fig:counterexample}, except for the fact that the number of degree-$3$ branches (i.e. the ones on the left side of $u$ in the figure) is now $7$ instead of $3$.

It only remains to analyze the distribution of $2$-dropouts. For this, note that the only information that $u$ can gather in $d=2$ rounds is the multiset of degrees of its neighbors. In practice, this will depend on the distance of the two removed nodes in the cycles; in particular, we can have the following cases:

1. If the two nodes are neighbors in (one of) the cycle(s), then due to the dropouts, $u$ will have two neighbors of degree $2$, and six neighbors of degree $3$. There are $2 \cdot \ell = 10$ possible cases to have this dropout combination in both graphs.

2. If the two nodes are at distance $2$ in (one of) the cycle(s), then $u$ will have a single neighbor of degree $1$, two neighbors of degree $2$ and five neighbors of degree $3$. This can again happen in $2 \cdot \ell = 10$ different ways in both graphs.

3. If the nodes have distance at least $3$ within the same cycle, or they are in different cycles, then the dropout creates four neighbors of degree $2$, and four neighbors of degree $3$. In the $2 \cdot \ell$ cycle, this can happen in $\frac{2 \cdot \ell \cdot (2 \cdot \ell - 5)}{2} = 2 \cdot \ell^2 - 5 \cdot \ell = 25$ different ways. In case of the two distinct $\ell$-cycles, this cannot happen in a single cycle at all (i.e. for general $\ell$, it can happen in $\frac{\ell \cdot (\ell - 5)}{2}$ ways, but this equals to $0$ for $\ell=5$); however, it can still happen if the two dropouts happen in different cycles, in $\ell \cdot \ell = 25$ different ways.

Hence the distribution of observed neighborhoods is also identical in case of $2$-dropouts.

\subsection{Proof of Theorem \ref{th:ports}}

The setting of Theorem \ref{th:ports} considers GNNs with port numbers (such as CPNGNN) where the neighborhood aggregation function is not permutation invariant, i.e. it can produce a different result for a different ordering of the inputs (neighbors) \citep{ports}. Our proof of the theorem already builds on the fact that one can extend the idea of injective GNNs (such as GIN in \citep{GIN}) to this setting with port numbers. To show that port numbers can be combined with the injective property, one can e.g.\ apply the same proof approach as in \citep{GIN}, using the fact that the possible combinations of embeddings and port numbers is still a countable set.

Given such an injective GNN with port numbers, the expressiveness of this GNN is once again identical to that of a general distributed algorithm in the message passing model with port numbers~\citep{ports}. As such, it suffices to show that a distributed algorithm in this model can separate any two different $d$-hop neighborhoods.

Let us assume the $1$-complete setting of Theorem \ref{th:chernoff1}, i.e. that we have sufficiently many runs to ensure that each $1$-dropout is observed at least once in the $d$-hop neighborhood of $u$. We show that the set of neighborhoods observed this way is sufficient to separate any two neighborhoods, regardless of the frequency of multi-dropout cases.

The general idea of the proof is that $1$-dropouts are already sufficient to recognize when two nodes in the tree representation of $u$'s neighborhood are actually corresponding to the same node. Consider three nodes $v_1$, $v_2$ and $v_3$, and assume that edges $(v_1, v_3)$ and $(v_2, v_3)$ are both within the $d$-hop neighborhood of $u$. More specifically, assume that $v_1$'s port number $b_1$ leads to $v_3$, and $v_2$'s port number $b_2$ also leads to $v_3$; then we can observe that the nodes at the endpoints of these two edges are always missing from the graph at the same time. That is, since we are guaranteed to observe every $1$-dropout at least once, if neighbor $b_1$ of $v_1$ and neighbor $b_2$ of $v_2$ are distinct nodes, then we must observe at least one neighborhood variant where only one of these two neighbors are missing; in this case, we know that the $b_1\,\!^{\text{th}}$ neighbor of $v_1$ and the $b_2\,\!^{\text{th}}$ neighbor of $v_2$ are not identical. On the other hand, if the two neighbors are always absent simultaneously, then the two edges lead to the same node.

The proof of the theorem happens in an inductive fashion. Note that from the $0$-dropout, we can already identify the degree of $u$ in the graph, and the port leading to each of its neighbors; this is exactly the $1$-hop neighborhood of $u$.

Now let us assume that we have already reconstructed the $(i-1)$-hop neighborhood of $u$; in this case, we can identify each outgoing edge from this neighborhood by a combination of a \textit{boundary node} (a node at distance $(i-1)$ from $u$) and a port number at this node. We can then extend our graph into the $i$-hop neighborhood of $u$ (for $i \leq d$) with the following two steps:

 1. First, we reconstruct the edges going from distance $(i-1)$ nodes to distance $i$ nodes. Let us refer to nodes at distance $i$ as \textit{outer} nodes. Note that all the outer neighbors of the boundary nodes can be identified by the specific outgoing edges from the boundary nodes; we only have to find out which of these outer nodes are actually the same. This can be done with the general idea outlined before: if two boundary nodes $v_1$ and $v_2$ have a neighbor at ports $b_1$ and $b_2$, respectively, and we do not observe a graph variant where only one of these neighbors is missing, then the two edges lead to the same outer node.

2. We also need to reconstruct the adjacencies between the boundary nodes; this is part of the $i$-hop neighborhood of $u$ by definition, but not part of the $(i-1)$-hop neighborhood. This happens with the same general idea as before: assume that $v_2$ and $v_3$ are both nodes at distance $(i-1)$, and $v_1$ is a node at distance $(i-2)$ that is adjacent to $v_3$. Then we can check whether $v_3$ disappears simultaneously from the respective ports $b_1$ and $b_2$ of nodes $v_1$ and $v_2$; if it does, then we know that edge $b_2$ of node $v_2$ leads to this other boundary node $v_3$.

After $d$ steps, this process allows us to reconstruct the entire $d$-hop neighborhood of $u$, thus proving the theorem.

Let us also briefly comment on the GNN interpretation of this graph algorithm. An injective GNN construction ensures that we map different $d$-hop neighborhoods to a different real number embedding. Note that the algorithm can separate any two neighborhoods without using the frequency of the specific neighborhoods variants; this implies that the set of real numbers obtained is different for any two neighborhoods, i.e. there must exist a number $z \in \mathbb{R}$ that is present in one of the distributions, but not in the other. One can then develop an MLP that essentially acts as an indicator for this value $z$, only outputting $1$ if the input is $z$; this allows us to separate the two neighborhoods.

Finally, note that our main objective throughout the paper was to compute a different embedding for two different neighborhoods. However, in this setting of Theorem \ref{th:ports}, it is also possible to encounter the opposite problem: if two $d$-hop neighborhoods are actually isomorphic, but they have a different assignment of port numbers, then they might produce a different embedding in the end.

We point out that with more sophisticated run aggregation, it is also possible to solve this problem, i.e. to recognize the same neighborhood regardless of the chosen port numbering. In particular, we have seen that in the $1$-complete case, the multiset of final embeddings already determines the entire neighborhood around $u$, and thus also its isomorphism class. This means that there is a well-defined function from the embedding vectors in $\mathbb{R}^r$ that we can obtain in $r$ runs to the possible isomorphism classes of $u$'s neighborhood (assuming for convenience that the neighborhood size is bounded). Due to the universal approximation theorem, a sufficiently complex MLP can indeed implement this function; as such, determining the isomorphism class of $u$'s neighborhood is indeed within the expressive capabilities of DropGNNs in this setting. However, while such a solution exists in theory, we note that this graph isomorphism problem is known to be rather challenging in practice.

\subsection{Briefly on the graph reconstruction problem}

The graph reconstruction problem is a well-known open question dating back to the 1940s. Assume that there is a hidden graph $G$ on $n \geq 3$ nodes that we are unaware of; instead, what we receive as an input is $n$ different modified variants of $G$, each obtained by removing a different node (and its incident edges) from $G$. This input multiset of graphs is often called the \textit{deck} of $G$. Note that the graphs in the deck are only provided up to an isomorphism class, i.e. for a specific node of the deck graph, we do not know which original node of $G$ it corresponds to. The goal is to identify $G$ from its deck; this problem is solvable exactly if there are no two non-isomorphic graphs with the same deck. This assumption is known as the graph reconstruction conjecture \citep{reconstruction}.

This problem is clearly close to our task of reconstructing a neighborhood from its $1$-dropout variants; however, there are also two key differences between the settings. Firstly, in our DropGNNs, we do not observe a graph, but rather a tree-representation of its neighborhood where some nodes may appear multiple times. In this sense, our GNN setting is much more challenging than the reconstruction problem, since it is highly non-trivial to decide whether two nodes in this tree representation correspond to the same original node. On the other hand, the DropGNN setting has the advantage that we can also observe the $0$-dropout; this does not happen in the reconstruction problem, since it would correspond to directly receiving the solution besides the deck.

\section{Dropouts with \texttt{mean} or \texttt{max} aggregation} \label{app:aggregate}

In this section, we discuss the expressiveness of the dropout technique with \texttt{mean} and \texttt{max} neighborhood aggregation. In particular, we prove that separation is always possible with \texttt{mean} aggregation when $|S_1| = |S_2|$, we construct a pair of neighborhoods that provide a very similar distribution of mean values, and we briefly discuss the limits of \texttt{max} aggregation in practice.

\subsection{Proof of Lemma \ref{th:mean}}

We begin with the proof of Lemma \ref{th:mean}. More specifically, we show that if $|S_1| = |S_2|$, then there always exists a choice of $p$ and integers $a, b$ such that after applying an activation function $\sigma(ax+b)$ on $S_1$ and $S_2$, a \texttt{mean} neighborhood aggregation allows us to distinguish the two sets.

In our proof, we assume that $S_1$ and $S_2$ are both multisets of integers (instead of vectors), i.e. that node features are only $1$-dimensional. With multi-dimensional feature vectors, we can apply the same proof to each dimension of the vectors individually; since $S_1 \neq S_2$, we will always have a dimension that allows us to separate the two multisets with the same method.

Let $\overline{s}_{1}$ denote the mean of $S_1$ and $\overline{s}_{2}$ denote the mean of $S_2$. We first discuss the simpler case when $\overline{s}_{1} \neq \overline{s}_{2}$; if this holds, we can distinguish any two sets $S_1$ and $S_2$, so we make this proof for the general case, without the assumption that $|S_1|=|S_2|$. After this, we discuss the case when $\overline{s}_{1} = \overline{s}_{2}$ and $|S_1|=|S_2|$; this completes the proof of Lemma \ref{th:mean}.

The main idea of the proofs is to find a threshold $\tau$ such that in $S_1$, we have mean values larger than $\tau$ much more frequently than in $S_2$ (or vice versa). We can then use an activation function $\hat{\sigma}(x):=\sigma(x-\tau)$ (with $\sigma$ denoting the Heaviside step function) to ensure that $\sigma(x)=1$ if $x \geq \tau$, and $\sigma(x)=0$ otherwise. This means that a run aggregation with \texttt{sum} will simply count the cases when the mean is larger than $\tau$, and thus with high probability, we get a significantly different sum in case of $S_1$ and $S_2$.

Note that even though the proof is described with a Heaviside activation function for ease of presentation, one could also use the logistic function (a more popular choice in practice), since the logistic function provides an arbitrary close approximation of the step function with the appropriate parameters.

\paragraph{When the means are different.}

First we consider the case when $\overline{s}_{1} \neq \overline{s}_{2}$.

In this setting, finding an appropriate $\tau$ is relatively straightforward. Assume w.l.o.g. that $\overline{s}_{1} < \overline{s}_{2}$, and let us choose an arbitrary $\tau$ such that $\overline{s}_{1} < \tau < \overline{s}_{2}$. This implies that whenever no node is removed, then the mean in $S_1$ will produce a $0$, while the mean in $S_2$ will produce a $1$.

It only remains to ensure that $0$-dropouts are frequent enough to distinguish these two cases. For this, let $\gamma = \max(|S_1|, |S_2|)$, and let us select $p=\frac{1}{2 \gamma}$. For both $S_1$ and $S_2$, this gives a probability of at least
\[ (1-p)^{\gamma} = \left( \frac{2 \gamma -1}{2 \gamma} \right)^{\gamma} \]
for $0$-dropouts. When $\gamma \geq 2$, this probability is strictly larger than $0.55$.

With a Chernoff bound, one can also show that the number of $0$-dropouts is strictly concentrated around this value: with $\delta=0.05$ and $r$ runs, the probability of the number of $0$-dropouts being below $(1-\delta) \cdot 0.55 \approx 0.52$ is upper bounded by $e^{-\frac{1}{3}\cdot \delta^2 \cdot 0.55 \cdot r}$. To ensure that this is below $\frac{1}{t}$, we only need $\Theta(1) \cdot r \geq \log t$, and hence $r \geq \Omega(\log t)$. This already ensures that in case of $S_2$, we have at least $0.52 \cdot r$ runs that produce a $1$, while in $S_1$, we have at least $0.52 \cdot r$ runs that produce a $0$ (i.e. at most $0.48 \cdot r$ runs that produce a $1$). Hence with high probability, a \texttt{sum} run aggregation gives a sum below $0.48 \cdot r$ and above $0.52 \cdot r$ for $S_1$ and $S_2$ respectively, so the two cases are indeed separable.

\paragraph{When the means are the same.}

Now consider the case when $\overline{s}_{1} = \overline{s}_{2}$, and we have $|S_1|=|S_2|$.

In this setting, let $\gamma=|S_1|=|S_2|$. Since the multisets are not identical, there must be an index $i \in \{ 1, ..., \gamma \}$ such that in the sorted version of the multisets, the $i^{\text{th}}$ element of $S_1$ is different from the $i^{\text{th}}$ element of $S_2$. Let us consider the smallest such index $i$, and assume w.l.o.g. that the $i^{\text{th}}$ element of $S_1$ (let us call it $x_{1, i}$) is larger than the $i^{\text{th}}$ element of $S_2$ (denoted by $x_{2, i}$). Furthermore, Let $\overline{s}_{1, -i}$ and $\overline{s}_{2, -i}$ denote the mean of $S_1$ and $S_2$, respectively, after removing the $i^{\text{th}}$ element.

Note that if we only had $1$-dropouts and $0$-dropouts in our GNNs, then finding this index $i$ would already allow a separation in a relatively straightforward way. Since $x_{1, i\!}>_{\!}x_{2, i}$, we must have $\overline{s}_{1, -i\!}<_{\!}\overline{s}_{2, -i}$. The idea is again to select a threshold value $\tau$ such that $\overline{s}_{1, -i}<\tau<\overline{s}_{2, -i}$. This ensures that in $S_1$, at least $i$ of the $1$-dropouts produce a $0$, whereas in $S_2$, at most $i-1$ of the $1$-dropouts produce a $0$. If the frequency of all $1$-dropouts is concentrated around its expectation, then this shows that the occurrences of $1$ will be significantly higher in $S_2$.

What makes this argument slightly more technical is the presence of $k$-dropouts for $k \geq 2$. In order to reduce the relevance of these cases, we select a smaller $p$ value. In particular, let $p=\frac{1}{2\gamma^2}$. In this case, the probability of a $k$-dropout is only
\[ p^k \cdot (1-p)^{\gamma-k} \leq p^k = \frac{1}{2^k \cdot \gamma^{2k}} \, ,\]
and the probability of having any multiple-dropout case in a specific run is at most
\[  \sum_{k=2}^{\gamma} \, {\binom{\gamma}{k}} \cdot \frac{1}{2^k \cdot \gamma^{2k}} \: \leq \: \sum_{k=2}^{\gamma} \, \frac{\gamma^k}{2} \cdot \frac{1}{2^k \cdot \gamma^{2k}} \: \leq \: \sum_{k=2}^{\gamma} \, \frac{1}{2^{k+1}} \cdot \frac{1}{\gamma^k} \: \leq \: \frac{1}{4 \cdot \gamma^2} \, , \]
using the fact that ${\binom{\gamma}{k}} \leq \frac{1}{2} \cdot \gamma^k$ for $k \geq 2$ and the fact that $\frac{1}{8}+\frac{1}{16}+... \leq \frac{1}{4}$.

On the other hand, the probability of a $1$-dropout is
\[ p \cdot (1-p)^{\gamma-1}  = \frac{1}{2 \gamma^2} \cdot \left( \frac{2 \gamma^2 -1}{2 \gamma^2} \right)^{\gamma-1} \, ,\]
where one can observe that the second factor is at least $\frac{7}{8}$ for any positive integer $\gamma$. As such, the probability of a $1$-dropout is lower bounded by $\frac{7}{16} \cdot \frac{1}{\gamma^2}$, i.e. it is notably larger than the cumulative probability of multiple-dropout cases.

This means that our previous choice of $\overline{s}_{1, -i}<\tau<\overline{s}_{2, -i}$ also suffices for this general case. In particular, even if all the multiple-dropouts in $S_1$ produce a mean that is larger than $\tau$, and all the multiple-dropouts in $S_2$ produce a mean that is smaller than $\tau$, we will still end up with a considerably larger probability of obtaining a value of $1$ in case of $S_2$, due to the $1$-dropout of the $i^{\text{th}}$ element. More specifically, the difference between the two probabilities will be at least $\frac{3}{16} \cdot \frac{1}{\gamma^2}$; using a Chernoff bound in a similar fashion to before, one can conclude that $\Omega(\gamma^4 \cdot \log t)$ runs are already sufficient to separate the two case with error probability at most $\frac{1}{t}$.

\subsection{Construction for similar mean distribution}

Let us now comment on the general case when we have $\overline{s}_{1} = \overline{s}_{2}$ but $|S_1| \neq |S_2|$. We present an example for two different sets $S_1$ and $S_2$ where the distribution of mean values obtained from $0$- and $1$-dropouts is essentially identical, thus showing the limits of any general approach that uses \texttt{mean} aggregation, but does not execute a deeper analysis of $k$-dropouts for $k \geq 2$.

Consider an even integer $\ell$, and consider the following two subsets. Let $S_1$ consist of $\frac{\ell}{2}$ distinct copies of the number $-(\ell-1)$, and $\frac{\ell}{2}$ distinct copies of the number $(\ell-1)$. Let $S_2$ consist of $\frac{\ell}{2}$ distinct copies of the number $-\ell$, and $\frac{\ell}{2}$ distinct copies of the number $\ell$, and a single instance of $0$. These sets provide $|S_1|=\ell$ and $|S_2|=\ell+1$, and also $\overline{s}_{1} = \overline{s}_{2}=0$. For a concrete example of $\ell=4$, we get the multisets $S_1=\{ -3,-3,3,3 \}$ and $S_2=\{ -4, -4, 0, 4, 4 \}$.

The mean values obtained for $1$-dropouts is also easy to compute in these examples. In $S_1$, we have $\frac{\ell}{2}$ distinct $1$-dropouts with a mean of $1$, and $\frac{\ell}{2}$ distinct $1$-dropouts with a mean of $-1$. In $S_2$, we have $\frac{\ell}{2}$ distinct $1$-dropouts with a mean of $1$, and $\frac{\ell}{2}$ distinct $1$-dropouts with a mean of $-1$, and a single $1$-dropout with a mean of $0$.

Note that if we only consider these $0$ and $1$-dropouts, then the probability of getting a $0$ is exactly the same in both settings. In $S_1$, this comes from the probability of the $0$-dropout only, so it is $(1-p)^{\ell}$. In $S_2$, we have to add up the probability of the $0$-dropout and a single $1$-dropout: this is $(1-p)^{\ell+1} + p \cdot (1-p)^{\ell} = (1-p)^{\ell}$.

The set of means obtained from $1$-dropouts is also identical in the two neighborhoods, it is only their probability that is slightly different. In $S_1$, both $-1$ and $1$ are obtained with probability $\frac{\ell}{2} \cdot p \cdot (1-p)^{\ell-1}$, while in $S_2$, they are both obtained with probability $\frac{\ell}{2} \cdot p \cdot (1-p)^{\ell}$. Hence the difference between the two probabilities is only
\[ \frac{\ell}{2} \cdot p \cdot \left((1-p)^{\ell-1} - (1-p)^{\ell}\right) = \frac{\ell}{2} \cdot p^2 \cdot (1-p)^{\ell-1} \, . \]
Recall that we have $\Theta(\ell^2)$ distinct $2$-dropouts, each with a probability of $p^2 \cdot (1-p)^{\ell-1}$, so these $2$-dropouts are together easily able to bridge this difference of frequency of the $1$-dropouts between $S_1$ and $S_2$. This shows that we cannot conveniently ignore multiple-node dropouts as in case of $|S_1|=|S_2|$ before: the only possible $1$-dropout-based approach to separate the two sets (i.e. to use the slightly different frequency of the values $-1$ and $1$) is not viable without a deeper analysis of the distributions of $2$-dropouts. It is beyond the scope of this paper to analyze this distribution in detail, or to come up with more sophisticated separation methods based on multiple-node dropouts.

\subsection{Aggregation with \texttt{max}}

Another well-known permutation-invariant function (and thus a natural candidate for neighborhood aggregation) is \texttt{max}; however, this method does not combine well with the dropout approach in practice.

In particular, if the multisets $S_1$ and $S_2$ only differ in their smallest element, then \texttt{max} aggregation can only distinguish them from a specific $(\gamma-1)$-dropout when all other neighbors of $u$ are removed. This dropout combination only has a probability of $p^{\gamma-1} \cdot (1-p)^2$; thus for a reasonably small $p$ (e.g. for $p \approx \gamma^{-1}$), we need a very high number of runs to observe this case with a decent probability.

\section{Details of the experimental setup}

In all of our experiments, we use Adam optimizer \citep{kingma2015adam}. For synthetic benchmarks and graph classification, we use a learning rate of $0.01$, for graph property regression we use a learning rate of $0.001$. For graph classification benchmarks we decay the learning rate by 0.5 every 50 steps \citep{GIN} and for the graph regression benchmark we decay the learning rate by a factor of $0.7$ on plateau \citep{morris2019weisfeiler}. The GIN model always uses 2-layer multilayer perceptrons and batch normalization \citep{ioffe2015batch} after each level \citep{GIN}. For our dropout technique, during preliminary experiments we tested three different node dropout implementation options: i) completely removing the dropped nodes and their edges from the graph; ii) replacing dropped node features by 0s before and after each graph convolution; iii) replacing the initial dropped node features by 0s. These preliminary experiments showed that all of these options performed similarly in practice, but the last option resulted in a more stable training. Since it is also the simplest dropout version to implement we chose to use it in all of our experiments. To ensure that the base model is well trained, when our technique is used we apply an auxiliary loss on each run individually. This auxiliary loss comprises $\frac{1}{3}$ of the final loss. While our model can have $O(n)$ memory consumption if we execute the runs in sequence, we implement it in a paralleled manner, which reduces the compute time, as all $r$ runs are performed in parallel, but increases memory consumption.

For the synthetic benchmarks (\textsc{Limits 1}, \textsc{Limits 2}, \textsc{4-Cycles}, \textsc{LCC}, \textsc{Triangles}, \textsc{Skip-Circles}) we use a GIN model with 4 convolutional layers (+ 1 input layer), \texttt{sum} as aggregation, $\varepsilon=0$ and for simplicity do not use dropout on the final \textsc{Readout} layer, while the final layer dropout is treated as a hyper-parameter in the original model. For synthetic node classification tasks (\textsc{Limits 1}, \textsc{Limits 2}, \textsc{LCC}, and \textsc{Triangles}) we use the same readout head as the original GIN model but skip the graph aggregation step. In all cases, except the \textsc{SkipCircles} dataset, 16 hidden units are used for synthetic tasks. For the \textsc{SkipCircles} dataset we use a GIN model with 9 convolutional layers (+ 1 input layer) with 32 hidden units as this dataset has cycles of up to 17 hops and requires long-range information propagation to solve the task. For the DropGIN variant, \texttt{mean} aggregation is used to aggregate node representations from different runs. When the GIN model is augmented with ports, which introduce edge features, we use modified GIN convolutions that include edge features \citep{hu2019strategies}. In synthetic benchmarks, we always generate the same number of graphs for training and test sets (generate a new copy of the dataset for testing) and for each random seed, we re-generate the datasets. We always feed in the whole dataset as one batch. \textsc{Limits 1}, \textsc{Limits 2} and \textsc{Skip-Circles} datasets are always comprised of graphs with the same structure, just with permuted node IDs for each dataset initialization, the remaining datasets have random graph structure, which changes when the datasets are regenerated. You can see the synthetic dataset structure type and statistics in Table~\ref{tab:synthetic_graphs}. All nodes in these datasets have the same degree.

\begin{table*}[ht]
\centering
\resizebox{0.9\textwidth}{!}{
\begin{tabular}{@{}l*{6}{S[table-format=-3.4]}@{}}
\toprule
{Dataset} & {Number of graphs} & {Number of nodes} & {Degree} & {Structure} & {Task} \\
\midrule
{\textsc{Limits 1} \citep{limits}} & \makebox{2} & \makebox{16} & \makebox{2} & \makebox{Fixed} & \makebox{Node classification}\\
{\textsc{Limits 2} \citep{limits}} & \makebox{2} & \makebox{16} & \makebox{3} & \makebox{Fixed} & \makebox{Node classification}\\
{\textsc{$4$-cycles} \citep{loukas2020graph}} & \makebox{50} & \makebox{16} & \makebox{2} & \makebox{Random} & \makebox{Graph classification}\\
{\textsc{LCC} \citep{randomFeatures1}} & \makebox{6} & \makebox{10} & \makebox{3} & \makebox{Random} & \makebox{Node classification}\\
{\textsc{Triangles} \citep{randomFeatures1}} & \makebox{1} & \makebox{60} & \makebox{3} & \makebox{Random} & \makebox{Node classification}\\
{\textsc{Skip-circles} \citep{chen2019equivalence}} & \makebox{10} & \makebox{41} & \makebox{4} & \makebox{Fixed} & \makebox{Graph classification}\\
\bottomrule
\end{tabular}}
\caption{Synthetic dataset statistics and properties.}  
\label{tab:synthetic_graphs}
\end{table*} 

For graph classification tasks we use exactly the same GIN model as described originally and apply our dropout technique on top. Namely, with 1 input layer, 4 convolution layers with \texttt{sum} as aggregation and $\varepsilon=0$ and dropout \citep{srivastava2014dropout} on the final \textsc{Readout} layer. For the DropGIN variant, \texttt{mean} aggregation is used to pool node representations from different runs. Note, that in our setting \texttt{sum} and \texttt{mean} aggregations are equivalent, up to a constant multiplicative factor, as the number of runs is a constant chosen on a per dataset level. We use exactly the same model training and selection procedure as described by \citep{GIN}. We decay the learning rate by 0.5 every 50 epochs and tune the number of hidden units $\in \{16, 32\}$ for bioinformatics datasets while using $64$ for the social graphs. The dropout ratio $\in \{0,0.5\}$ after the final dense layer the batch size $\in \{32,128\}$ are also tuned. The epoch with the best cross-validation accuracy over the 10 folds is selected. You can see the statistics of synthetic datasets in Table~\ref{tab:real_graphs}.

\begin{table*}[ht]
\centering
\resizebox{0.65\textwidth}{!}{
\begin{tabular}{@{}l*{8}{S[table-format=-3.4]}@{}}
\toprule
 &  & \multicolumn{3}{c}{Number of nodes} & \multicolumn{3}{c}{Degree} \\
\cmidrule(lr){3-5} \cmidrule(lr){6-8}
{Dataset} & {Number of graphs} &{Min} & {Max} & {Mean} & {Min} & {Max} & {Mean}\\
\midrule
{\textsc{MUTAG}} & \makebox{188} & \makebox{10} & \makebox{28} & \makebox{18} & \makebox{3} & \makebox{4} & \makebox{3.01} \\
{\textsc{PTC}} & \makebox{344} & \makebox{2} & \makebox{64} & \makebox{14} & \makebox{1} & \makebox{4} & \makebox{3.18} \\
{\textsc{Proteins}} & \makebox{1109} & \makebox{4} & \makebox{336} & \makebox{38} & \makebox{3} & \makebox{12} & \makebox{5.78}\\
{\textsc{IMDB-B}} & \makebox{996} & \makebox{12} & \makebox{69} & \makebox{19} & \makebox{11} & \makebox{68} & \makebox{18.49}\\
{\textsc{IMDB-M}} & \makebox{1498} & \makebox{7} & \makebox{63} & \makebox{13} & \makebox{6} & \makebox{62} & \makebox{11.91}\\
\midrule
{\textsc{QM9}} & \makebox{130 831} & \makebox{3} & \makebox{29} & \makebox{18} & \makebox{2} & \makebox{5} & \makebox{3.97}\\
\bottomrule
\end{tabular}}
\caption{Real-world dataset statistics.}  
\label{tab:real_graphs}
\end{table*} 

For the graph property regression task (\textsc{QM9}) we augment two models: 1-GNN \citep{morris2019weisfeiler} and MPNN \citep{gilmer2017neural}. For 1-GNN we use the code and the training setup as provided by the original authors\footnote{\url{https://github.com/chrsmrrs/k-gnn}}. For MPNN we use the reference model implementation from PyTorch Geometric \footnote{\url{https://github.com/rusty1s/pytorch_geometric/blob/master/examples/qm9_nn_conv.py}}. We otherwise follow the training and evaluation procedure used by 1-GNN \citep{morris2019weisfeiler}. The models are trained for 300 epochs and the epoch with the best validation score is chosen. 

We use PyTorch \citep{paszke2019pytorch} and PyTorch Geometric \citep{fey2019fast} for the implementation. All models have been trained on Nvidia Titan RTX GPU (24GB RAM).

\end{document}